\documentclass[pdflatex,sn-mathphys-num]{sn-jnl}% Math and Physical Sciences Numbered Reference Style
\usepackage{graphicx}%
\usepackage{multirow}%
\usepackage{amsmath,amssymb,amsfonts}%
\usepackage{amsthm}%
\usepackage{mathrsfs}%
\usepackage[title]{appendix}%
\usepackage{xcolor}%
\usepackage{textcomp}%
\usepackage{manyfoot}%
\usepackage{booktabs}
\usepackage{multirow}
\usepackage{algorithm}%
\usepackage{algorithmicx}%
\usepackage{algpseudocode}%
\usepackage{listings}%
\PassOptionsToPackage{hyphens}{url}\usepackage{hyperref}
\usepackage{float}
\usepackage{tabularx,booktabs,ragged2e}
\usepackage{array}
\usepackage{subfigure}
\usepackage{longtable}
\usepackage{natbib}
\usepackage{adjustbox}
\usepackage{url}
\usepackage{colortbl}
\usepackage{multirow}
\usepackage{titlesec}
\usepackage{comment}
\usepackage{xcolor}
\usepackage{caption}
\usepackage{makecell}
\usepackage{subcaption}

\theoremstyle{thmstyleone}%
\theoremstyle{thmstyletwo}%

\theoremstyle{thmstylethree}%

\begin{document}

\title[Article Title]{A Model Ensemble-Based Post-Processing Framework for Fairness-Aware Prediction}


\author*[1]{\fnm{Zhouting} \sur{Zhao}}\email{zhaozh@tcd.ie}

\author[1]{\fnm{Tin Lok James} \sur{Ng}}\email{ngja@tcd.ie}

\affil*[1]{\orgdiv{School of Computer Science and Statistics}, \orgname{Trinity College Dublin}, \orgaddress{\city{Dublin}, \country{Ireland}}}


\abstract{
Striking an optimal balance between predictive performance and fairness continues to be a fundamental challenge in machine learning. In this work, we propose a post-processing framework that facilitates fairness-aware prediction by leveraging model ensembling. Designed to operate independently of specific model internals, our approach can be instantiated across different learning tasks and model architectures using task-specific fairness definitions. Through experiments spanning classification, regression, and survival analysis, we show that the framework provides adjustable fairness--performance trade-offs and can reduce group-fairness disparities in a range of experimental settings.
}




\maketitle

\section{Introduction}
As machine learning systems 
increasingly influence decisions in critical domains, ensuring their fairness has become a vital concern \cite{barocas2020fairness}. Algorithmic models trained on historical or biased data risk perpetuating or amplifying existing inequalities, raising ethical, social, and legal challenges. Fairness considerations are particularly important across various application domains. For example, classification models are widely used in loan approvals, hiring decisions, and criminal justice risk assessments. Regression models inform salary predictions and insurance pricing. Clustering techniques play a key role in customer segmentation and healthcare group analysis, while survival analysis is crucial for estimating patient survival times or employee turnover. 

In simple terms, fairness in machine learning refers to the idea that a model’s predictions should not systematically favor or disadvantage certain individuals or groups defined by sensitive attributes such as gender, race, or age. Fairness can be measured using various quantitative criteria that assess disparities in model outcomes between these groups (e.g., differences in predicted probabilities, false positive rates, or expected values). The specific notion of fairness considered often depends on the task and the ethical or policy context of the application.

Various fairness metrics and strategies to mitigate unfairness have been proposed across multiple tasks, including classification \cite{zafar2017fairness, dwork2012fairness, kamishima2012fairness}, regression \cite{filippi2023local, berk2017convex, agarwal2019fair}, clustering \cite{chierichetti2017fair, kleindessner2020notion, li2021approximate}, and survival analysis \cite{sonabend2022flexible, keya2021equitable, rahman2022fair, zhao2023fairness}. Strategies for mitigating unfairness can be broadly categorized into three main approaches: pre-processing, in-processing, and post-processing \cite{pessach2022review, mehrabi2021survey, caton2024fairness}, each with its own advantages and limitations. Pre-processing methods modify the input data to reduce bias before model training; in-processing approaches intervene during model training by incorporating fairness constraints or regularization terms; and post-processing methods adjust the model outputs to improve fairness after training.

The key advantage of pre-processing methods, such as reweighing \cite{kamiran2012data} and disparate impact remover \cite{feldman2015certifying}, lies in their model-agnostic nature, as they function independently of the training process and can be easily applied to various algorithms. However, modifying the dataset to remove biases can lead to a loss of valuable information, potentially reducing model performance. Additionally, such modifications may introduce unintended biases or distort the original data distribution, which can negatively impact downstream tasks.

In-processing techniques embed fairness objectives directly into model optimization by incorporating additional fairness constraints or regularization terms \cite{zhang2018mitigating,kamishima2012fairness,han2023ffb}. While these methods can effectively balance fairness and predictive performance, they often lead to increased computational complexity. Unlike pre-processing methods, in-processing approaches are typically model-specific, meaning a fairness intervention designed for one model architecture may not easily transfer to another, limiting flexibility. 

Post-processing methods for fairness are techniques applied after a machine learning model has been trained and its predictions generated \cite{pleiss2017fairness}. These methods adjust the outputs to meet specific fairness criteria without modifying the model or the training data. However, many post-processing approaches are tailored to specific tasks (e.g., binary classification) or fairness notions (e.g., demographic parity). To overcome these limitations, Tifrea et al.~\cite{tifreafrappe} proposed \textsc{FRAPPÉ} (Fairness Framework for Post-Processing Everything), a general post-processing framework that transforms fairness-aware in-processing methods into post-processing ones. By introducing an additive correction term, \textsc{FRAPPÉ} supports broader fairness constraints and is applicable to both classification and regression. Another benefit of \textsc{FRAPPÉ} is that it does not rely on sensitive attributes during inference.

Despite the notable advantages of FRAPPÉ compared to other post-processing fairness methods, its additive correction formulation represents one particular mechanism for adjusting the outputs of a trained model. In this work, we explore an alternative based on model ensembling: specifically, we consider taking a weighted average of a pre-trained complex model and a simpler model. This approach blends the predictions from both models, indirectly promoting fairness by tuning the ensemble weights and the parameters of the simpler model, while keeping the original trained model unchanged.

Moreover, while FRAPPÉ is suitable for binary classification and regression tasks, extending it to the increasingly important domain of survival analysis (also known as time-to-event prediction) is non-trivial. Survival analysis has become a critical area in machine learning, especially in fields like healthcare, finance, and engineering, where predicting when an event will occur is as important as predicting if it will occur. Unlike standard classification or regression, fairness in survival analysis often involves maintaining parity across demographic groups over time or balancing expected survival probabilities, goals that are not easily addressed through FRAPPÉ’s current additive framework. In contrast, the proposed ensemble-based formulation can be instantiated for survival analysis by combining task-appropriate survival predictions and fairness criteria within the same post-processing framework.

Our proposed ensemble approach builds on the principles of mixture modeling and the mixture-of-experts paradigm \cite{jacobs1991adaptive}, where the former uses fixed mixture weights and the latter allows weights to vary with input features. In our framework, the pre-trained model remains fixed, and the learning process focuses solely on estimating the mixture weights and the parameters of a simpler auxiliary model. This design restricts the post-processing step to estimating the mixture weights and, where applicable, the parameters of a relatively simple auxiliary model, while keeping the original trained model fixed. Like FRAPPÉ, our method is model-agnostic, supports a wide range of fairness definitions, and does not require access to sensitive attributes at inference time. Furthermore, our experimental results illustrate the range of fairness--performance trade-offs obtained under different datasets and model configurations.

In certain scenarios, two separate pre-trained models may be available: one optimized for predictive performance and another designed to promote fairness. The fairness-oriented model might be trained on pre-processed inputs or built using a simpler model with a deliberately restricted set of non-sensitive features. Our proposed ensemble method can be adapted to this setting, where both models are pre-trained. The objective is to combine their predictions and characterize the resulting fairness--performance trade-off. In this variant, the ensemble step consists solely of learning the combination weights, with both component models remaining fixed. 

In summary, our contributions are as follows:
\begin{enumerate}
\item We introduce a post-processing framework for fairness-aware prediction based on model ensembling, which is model-agnostic with respect to the internal structure of the pre-trained performance-oriented model.
\item We develop two complementary ensemble formulations: a global Mixture model and an input-dependent MoE model. These formulations support scenarios in which the fairness-oriented component is either learned during post-processing or provided as an independently pre-trained model.
\item We instantiate and evaluate the framework for binary classification, regression, and survival analysis using eight real-world datasets. The results demonstrate that the proposed configurations can reduce group-fairness disparities while offering adjustable trade-offs with predictive performance.
\end{enumerate}

\section{Background and Related Work}

\subsection{Post-processing for Fairness}
Post-processing techniques modify model predictions after training and are particularly valuable when working with pre-trained or black-box models. These approaches are model-agnostic and well-suited for deployment. A variety of post-processing methods have been developed. For instance, Pleiss et al.~\cite{pleiss2017fairness} proposed a calibrated post-processing approach that adjusts predictions to improve group fairness while accounting for calibration. Petersen et al.~\cite{petersen2021post} introduced a graph-based post-processing method for individual fairness based on graph Laplacian smoothing. Di Gennaro et al.~\cite{di2024post} proposed a model-agnostic post-processing framework that reduces unfairness while limiting the deviation between the original and adjusted predictions. Xian et al.~\cite{xian2023fair} employed a Wasserstein barycenter framework to construct fair classifiers from score functions under demographic parity constraints. Jiang et al.~\cite{jiang2020wasserstein} used Wasserstein-1 distance to derive fair post-processed classifiers with improved robustness to threshold selection. In the multi-class setting, Alghamdi et al.~\cite{alghamdi2022beyond} developed a projection-based technique that maps the outputs of a pre-trained classifier onto the space of fair predictors. Despite these developments, most post-processing approaches remain focused primarily on classification and are often designed around particular fairness criteria, which can limit their applicability across prediction tasks and fairness objectives.
\\\\
To address this limitation, Tifrea et al.~\cite{tifreafrappe} proposed FRAPPÉ, a general and model-agnostic post-processing framework. 
To make the discussion concrete, we focus on the regression setting. Given a pre-trained model \( f_{\text{base}} \), the FRAPPÉ framework introduces an additive linear perturbation to its output:
\begin{equation}
        f_{\text{comb}}(\mathbf{x}) = f_{\text{base}}(\mathbf{x}) + T_{\text{PP}}(\mathbf{x})
\label{eq:frappe_output}
\end{equation}
where $\mathbf{x}$ represents the features and \( T_{\text{PP}}(\mathbf{x}) \) is defined as a linear function of features:
 
\begin{equation}
    T_{\text{PP}}(\mathbf{x}) = \boldsymbol{\theta}^\top \mathbf{x}.
\label{eq:linear_perturbation}
\end{equation}
The optimization objective is defined as:
\begin{equation}
    \text{arg} \min_{T_{\text{PP}}} \frac{1}{|{\cal D}_{\text{pred}}|} \sum_{\mathbf{x} \in \mathcal{D}_{\text{pred}}} d_{\text{pred}}(f_{\text{comb}}(\mathbf{x}), f_{\text{base}}(\mathbf{x})) + \lambda \, \mathcal{L}_{\text{fair}}(f_{\text{comb}}; \mathcal{D}_{\text{sensitive}}),
\label{eq:frappe_obj}
\end{equation}
where \( d_{\text{pred}} \) measures dissimilarity in the prediction space, and \( \mathcal{L}_{\text{fair}} \) is a fairness regularization term. When $T_\text{PP}$ is a linear function \eqref{eq:linear_perturbation}, the minimization in \eqref{eq:frappe_obj} is performed with respect to the parameter $\boldsymbol{\theta}$. The datasets \( \mathcal{D}_{\text{pred}} = \{\mathbf{x}_i\}_{i=1}^{n_1} \) and \( \mathcal{D}_{\text{sensitive}} = \{(\mathbf{x}'_j, y'_j, \mathbf{a}'_j)\}_{j=1}^{n_2} \) in \eqref{eq:frappe_obj} are assumed to be drawn from the same distribution and may overlap. The first term in the objective function is designed to ensure that the perturbation \( T_{\text{PP}} \) does not significantly alter the predictions of the base model. The parameter \( \lambda > 0 \) governs the trade-off between fidelity to the base model and fairness.

\subsection{Fairness Metrics}

Fairness metrics provide formal criteria to assess whether machine learning models produce equitable predictions across subpopulations defined by sensitive attributes. These metrics are typically categorized into \textit{group fairness} and \textit{individual fairness} \cite{caton2024fairness,pessach2022review,mehrabi2021survey}. Group fairness aims to ensure that outcomes are distributed equitably across predefined demographic groups, promoting parity regardless of group membership. In contrast, individual fairness emphasizes treating similar individuals in a similar manner, which typically relies on task-specific similarity metrics.

In this work, we focus on group fairness metrics and adopt task-specific measures aligned with established literature. For binary classification, we utilize Demographic Parity \cite{calders2009building}, as defined in Equation \eqref{eq:binary_fairness_metric}, which requires that the probability of a positive prediction be independent of group membership. For regression tasks, we adopt Statistical Parity AUC \cite{filippi2023local}, as defined in Equation \eqref{eq:regression_fairness_metric}, which quantifies disparities in predicted outcomes between groups across the full range of output thresholds. In the context of survival analysis, we employ the concept of Group Fairness \cite{keya2021equitable, rahman2022fair}, as defined in Equation \eqref{eq:survival_group_fairness_metric}, which evaluates the maximum deviation between group-specific and overall expected survival probabilities over time. While the empirical evaluation focuses on these task-specific group-fairness metrics, the framework permits alternative fairness regularization terms when they can be expressed within the proposed optimization objective.

\subsection{Mixture of Experts}
Our proposed ensemble approach to fairness is inspired by the Mixture-of-Experts (MoE) framework \cite{jacobs1991adaptive}. MoE is a machine learning paradigm that decomposes a complex learning task into multiple specialized sub-models, or “experts,” each tailored to a specific region of the input space. These experts are combined through a gating mechanism that adaptively assigns weights to each expert based on the input features. This structure is especially well-suited for capturing heterogeneous data distributions, as it enables model components to specialize while maintaining strong overall predictive performance. During training, the expert models and the gating network are jointly optimized to minimize a unified objective function.

Formally, given input feature \( \mathbf{x} \in \mathcal{X} \), the MoE model is defined as:

\begin{equation}
f_{\text{MoE}}(\mathbf{x}) = \sum_{i=1}^K \alpha_i(\mathbf{x}; \phi) \cdot f_i(\mathbf{x}; \theta_i),
\label{eq:moe}
\end{equation}

where \( f_i(\mathbf{x}; \theta_i) \) is the \( i \)-th expert model with learnable parameters \( \theta_i \), and \( \alpha_i(\mathbf{x}; \phi) \in [0, 1] \) is the gating weight computed by a gating network parameterized by \( \phi \). The weights are constrained such that \( \sum_{i=1}^K \alpha_i(\mathbf{x}; \phi) = 1 \), typically implemented using a softmax function over transformed inputs. This formulation enables both the model architecture and the instance-specific routing strategy to be optimized simultaneously. Recently, MoE has also been applied to fairness-aware learning, for example, FairMOE \cite{kamishima2024fairmoe} enforces counterfactual fairness by limiting the influence of sensitive attributes on expert routing.

%Recent MoE research has focused on scaling to large neural architectures by activating only a sparse subset of experts per input \cite{shazeer2017outrageously, fedus2022switch}, with success in areas such as machine translation and language modeling \cite{lepikhin2020gshard, du2022glam}. MoE has also been extended to fairness-aware learning, for example, FairMOE \cite{kamishima2024fairmoe} enforces counterfactual fairness by limiting the influence of sensitive attributes on expert routing. Compared to traditional ensemble methods such as bagging, boosting, and stacking \cite{dietterich2000ensemble, opitz1999popular}, which combine independently trained models using fixed rules, MoE learns instance-dependent weights jointly with expert models. This allows finer control but often requires end-to-end training. Our framework combines elements of both: we use separately trained expert models and learn a post-hoc merging function to balance fairness and performance. 

\section{An Ensemble Approach to Fairness, One Pre-trained Model}
\label{sec_method_one_model}

\subsection{General setting}
\label{sec_general_framework}
We consider a setting in which a pre-trained model, optimized for predictive performance, is available. Our goal is to post-process this model to enhance its fairness. Like FRAPPÉ, our approach involves modifying the predictions of the pre-trained model to achieve this objective. However, rather than applying a direct perturbation to the model’s outputs, we adopt an alternative strategy by combining the pre-trained model with a simpler model through a MoE framework. In this formulation, the pre-trained model remains fixed, while we jointly learn the parameters of the simpler model and the input-dependent gating weights that determine the contribution of each model in the ensemble.

We begin by considering a general setting, where \(\mathcal{X}\) denotes the feature space and \(\mathcal{Y}\) represents the output space, which varies depending on the task. For binary classification, \(\mathcal{Y}=[0,1]\); for regression, \(\mathcal{Y} \subseteq \mathbb{R}\); and for survival analysis, \(\mathcal{Y}\) corresponds to the set of survival functions
\begin{eqnarray*}
\mathcal{Y} = \Big\{ S: \mathcal{X} \times \mathbb{R}_+ \to [0,1] \;\Big|\;
\forall \mathbf{x} \in \mathcal{X}, \; S(\cdot \mid \mathbf{x}) \text{ is non-increasing, } \\ S(0 \mid \mathbf{x}) = 1, \; \lim_{t \to \infty} S(t \mid \mathbf{x}) = 0 \Big\}.    
\end{eqnarray*}
We assume the availability of a pre-trained model \( f_{\text{perf}}: \mathcal{X} \rightarrow \mathcal{Y} \), whose internal architecture, training procedure, and training dataset are not accessible. Consequently, the model is treated as a black-box. Our post-processing approach introduces a simpler model \( f_{\text{simp}}: \mathcal{X} \rightarrow \mathcal{Y} \) and a gating function \( \alpha: \mathcal{X} \rightarrow \mathbb{R} \). We jointly learn the model \( f_{\text{simp}} \) and the gating function \( \alpha \).

Unlike the standard MoE setting, however, the pre-trained model \( f_{\text{perf}} \) remains fixed during the training process.
The combination of the pre-trained model and the simpler model yields the following weighted prediction function:
\begin{eqnarray}
\label{eqn_weighted_model}
f_{\text{combined}}(\mathbf{x}) = \alpha(\mathbf{x}) f_{\text{perf}}(\mathbf{x}) + (1 - \alpha(\mathbf{x})) f_{\text{simp}}(\mathbf{x}),    
\end{eqnarray}
where \(\alpha(\mathbf{x})\) denotes the input-dependent gating weight. We note that our MoE setup involves two models: a fixed model $f_{\text{perf}}$ and a trainable model $f_{\text{simp}}$, representing a special case of a two-expert MoE. For notational simplicity, we suppress the explicit parameter dependencies of the gating function \(\alpha\) and the simpler model \(f_{\text{simp}}\).

Our optimization objective follows the same formulation as FRAPPÉ and is given by:
\begin{eqnarray}
\label{eqn_MOE_obj}
\min_{\alpha, f_{\text{simp}}} 
\frac{1}{|{\cal D}_{\text{pred}}|} \sum_{\mathbf{x} \in \mathcal{D}_{\text{pred}}} 
d_{\text{pred}}(f_{\text{combined}}(\mathbf{x}), f_{\text{perf}}(\mathbf{x})) + 
\lambda \, \mathcal{L}_{\text{fair}}(f_{\text{combined}}; \mathcal{D}_{\text{sensitive}}),    
\end{eqnarray}
where \(d_{\text{pred}}\) measures predictive deviation from the pre-trained model, \(\mathcal{L}_{\text{fair}}\) denotes a fairness regularization term, \( \mathcal{D}_{\text{pred}} = \{\mathbf{x}_i\}_{i=1}^{n_1} \) and \( \mathcal{D}_{\text{sensitive}} = \{(\mathbf{x}'_j, y'_j, \mathbf{a}'_j)\}_{j=1}^{n_2} \). The trade-off between fidelity (i.e., maintaining consistency with its predictions) to the original model and fairness is controlled by the parameter \(\lambda\).

Many choices for gating weight \(\alpha(\mathbf{x})\) are possible. We consider the following logistic formulation:
\begin{eqnarray}
\label{eqn_gating_weights}
\alpha(\mathbf{x}) = \frac{1}{1 + \exp(-(\mathbf{x} \cdot \boldsymbol{\beta} + \beta_0))},    
\end{eqnarray}
where \(\boldsymbol{\beta}\) are the coefficients associated with the features in \(\mathbf{x}\), and \(\beta_0\) is an intercept term. 

The choice of the simpler model \(f_{\text{simp}}\) is task-dependent. In principle, different task-appropriate model classes may be incorporated, provided that their outputs are compatible with the combination rule and the resulting objective can be optimized. For binary classification, a common choice is \(f_{\text{simp}}(\mathbf{x}) = \sigma(\mathbf{x} \cdot \boldsymbol{\gamma})\), where \(\sigma(\cdot)\) denotes the sigmoid function and $\boldsymbol{\gamma}$ represents the model parameters. For regression tasks, we may use a linear model, \(f_{\text{simp}}(\mathbf{x}) = \mathbf{x} \cdot \boldsymbol{\gamma}\). In the context of survival analysis, $f_{\text{simp}}(\mathbf{x})$ can be defined as the survival function derived from the Cox proportional hazards model, assuming a constant baseline hazard. Specifically, assuming a constant baseline hazard equal to 1, the hazard function is $\lambda(t \mid \mathbf{x}) = \exp(\mathbf{x} \cdot \boldsymbol{\gamma}),$
and the corresponding survival function is $S(t \mid \mathbf{x}) = \exp\big(- t \, \exp(\mathbf{x} \cdot \boldsymbol{\gamma})\big).$
\\\\
\textbf{Comparison with \textsc{FRAPPÉ}.}
\\\\
A comparison between the proposed ensemble-based post-processing approach and \textsc{FRAPPÉ} highlights both conceptual similarities and important structural differences. Both methods leave the original pre-trained model unchanged and optimize a post-processing objective that balances fidelity to the original predictions against a fairness regularization term. The main distinction lies in the parameterization of the post-processing adjustment. As introduced in Section~2.1, \textsc{FRAPPÉ} modifies the output of the pre-trained model through the additive perturbation in Equation~\eqref{eq:frappe_output}, whereas our framework forms the adjusted prediction through the weighted combination in Equation~\eqref{eqn_weighted_model}.

More formally, Equation~\eqref{eqn_weighted_model} can be rewritten as
\begin{equation}
\label{eq:ensemble_correction}
f_{\text{combined}}(\mathbf{x})
=
f_{\text{perf}}(\mathbf{x})
+
\bigl(1-\alpha(\mathbf{x})\bigr)
\left(
f_{\text{simp}}(\mathbf{x})
-
f_{\text{perf}}(\mathbf{x})
\right).
\end{equation}
Thus, the proposed approach can also be viewed as applying a correction to the original prediction, but the correction is structurally different from that used in \textsc{FRAPPÉ}. Rather than learning an additive perturbation directly, the correction is anchored to the difference between two task-compatible predictors and its magnitude is controlled by the gating weight. In this sense, the post-processing adjustment is determined by how much the prediction is moved from the performance-oriented model toward the simpler fairness-oriented component.

This structure also constrains the set of attainable post-processed predictions. Since \(\alpha(\mathbf{x})\in[0,1]\), Equation~\eqref{eqn_weighted_model} implies
\begin{equation}
\label{eq:convex_hull_property}
f_{\text{combined}}(\mathbf{x})
\in
\operatorname{conv}
\left\{
f_{\text{perf}}(\mathbf{x}),
f_{\text{simp}}(\mathbf{x})
\right\}.
\end{equation}
For scalar predictions, this is equivalent to
\[
\min\left\{
f_{\text{perf}}(\mathbf{x}),
f_{\text{simp}}(\mathbf{x})
\right\}
\leq
f_{\text{combined}}(\mathbf{x})
\leq
\max\left\{
f_{\text{perf}}(\mathbf{x}),
f_{\text{simp}}(\mathbf{x})
\right\}.
\]
The proposed post-processing mechanism therefore interpolates between two predictive models, whereas the linear additive perturbation used in \textsc{FRAPPÉ} is not, in general, constrained to remain between the original prediction and the output of a second predictive model.

The two approaches consequently induce different post-processing function classes. In the formulation considered in Section~2.1, \textsc{FRAPPÉ} uses a linear perturbation \(T_{\text{PP}}(\mathbf{x})=\boldsymbol{\theta}^{\top}\mathbf{x}\). By contrast, the correction induced by Equation~\eqref{eq:ensemble_correction} is
\[
\bigl(1-\alpha(\mathbf{x})\bigr)
\left(
f_{\text{simp}}(\mathbf{x})
-
f_{\text{perf}}(\mathbf{x})
\right),
\]
which is generally nonlinear whenever the performance-oriented model is nonlinear, even when the simpler model itself is linear. In the MoE formulation, the input-dependent gating function further allows the extent of this adjustment to vary across observations. Hence, the proposed ensemble formulation is not merely a reparameterization of the linear perturbation used in \textsc{FRAPPÉ}, but defines a structurally different family of post-processing transformations.

These structural differences have further task-specific implications. In particular, the convex-combination formulation preserves valid probability outputs in classification and provides a natural mechanism for combining structured survival predictions, as discussed in Section~\ref{subsec:applications}.
\\\\
\textbf{Mixture Modeling}
\\\\
In scenarios where the gating weight function \(\alpha(\mathbf{x})\) exhibits minimal variability across the input space, it may be reasonable to adopt a simplified modeling approach by treating \(\alpha(\mathbf{x})\) as a constant. This simplification effectively reduces the model from a full MoE formulation to a standard mixture model. In this setting, the combination rule for predictions becomes static, and the weighted prediction function in Equation~\eqref{eqn_weighted_model} simplifies to:

\begin{equation}
\label{eqn_weighted_model_mixture}
f_{\text{combined}}(\mathbf{x}) = \alpha f_{\text{perf}}(\mathbf{x}) + (1 - \alpha) f_{\text{simp}}(\mathbf{x}),
\end{equation}

\noindent
where \(\alpha \in [0, 1]\) is a fixed mixing coefficient that governs the trade-off between the pre-trained performance-oriented model \(f_{\text{perf}}\) and the simpler model \(f_{\text{simp}}\). This approach is particularly suitable in settings where fairness adjustments are meant to be applied uniformly across the entire population. 

In addition to providing a simpler parameterization, the global Mixture formulation gives the mixing coefficient a direct interpretation in terms of deviation from the performance-oriented model. From Equation~\eqref{eqn_weighted_model_mixture},
\begin{equation}
\label{eq:mixture_deviation}
f_{\text{combined}}(\mathbf{x}) - f_{\text{perf}}(\mathbf{x})
=
(1-\alpha)
\left(
f_{\text{simp}}(\mathbf{x}) - f_{\text{perf}}(\mathbf{x})
\right).
\end{equation}
Hence, \(\alpha=1\) exactly recovers the original performance-oriented prediction, whereas decreasing \(\alpha\) moves the prediction toward the simpler component. The global Mixture therefore provides an explicit and interpretable interpolation path between the two component models.

\subsection{Construction of the Simpler Model}
\label{sec:simpler_model_construction}

The simpler model \(f_{\mathrm{simp}}\) is intended to provide a low-complexity, fairness-oriented component that complements \(f_{\mathrm{perf}}\). Its input representation is therefore intentionally restricted to a small subset of non-sensitive features rather than the full feature set available to \(f_{\mathrm{perf}}\). This restriction preserves the intended simplicity of \(f_{\mathrm{simp}}\) and limits the amount of information available to the simpler model when forming its predictions.

Let \(\mathcal{F}\) denote the set of available non-sensitive features and let \(S \subseteq \mathcal{F}\) be a candidate feature subset. For each \(S\), let \(f_S\) denote the corresponding task-specific simpler model fitted using only the features in \(S\). Feature selection is performed using the training data only, and all fairness and predictive-loss quantities appearing below are evaluated on the training data.

A candidate simpler model is considered fairness-admissible if
\begin{equation}
    \mathcal{L}_{\mathrm{fair}}(f_S)
    \leq
    \epsilon_D,
    \label{eq:fair_model_admissibility}
\end{equation}
where \(\epsilon_D \geq 0\) is a dataset- and task-specific tolerance defining the maximum admissible disparity. This allows the admissibility criterion to account for differences in the empirical scale of the fairness measure across datasets and prediction tasks. The value of \(\epsilon_D\) is fixed before the subsequent post-processing optimization.

The preferred feature cardinality is defined as
\begin{equation}
    k^{*}
    =
    \min
    \left\{
        |S| :
        S \subseteq \mathcal{F},
        \;
        \mathcal{L}_{\mathrm{fair}}(f_S)
        \leq
        \epsilon_D
    \right\}.
    \label{eq:min_cardinality_simple_model}
\end{equation}

Candidate subsets are considered in increasing order of cardinality, so that higher-dimensional subsets are considered only when no lower-dimensional subset satisfies the fairness-admissibility criterion. If multiple subsets are fairness-admissible at the minimum cardinality \(k^{*}\), predictive utility is used as a secondary selection criterion. The final subset is therefore selected as
\begin{equation}
    S^{*}
    =
    \arg\min_{
        \substack{
            S \subseteq \mathcal{F},\\
            |S| = k^{*},\\
            \mathcal{L}_{\mathrm{fair}}(f_S)
            \leq
            \epsilon_D
        }
    }
    \mathcal{L}_{\mathrm{pred}}(f_S),
    \label{eq:simple_model_selection}
\end{equation}
where \(\mathcal{L}_{\mathrm{pred}}\) denotes the task-specific predictive loss.

This two-stage construction prioritizes low model complexity and fairness admissibility, while avoiding the selection of an unnecessarily weak simpler model when several candidates satisfy the fairness requirement.

\subsection{Applications}
\label{subsec:applications}
We now illustrate the application of the general framework introduced in Section~\ref{sec_general_framework} across three distinct prediction tasks: binary classification, regression, and survival analysis. The formulation below accommodates both the MoE scenario, where the weighting function \( \alpha(\mathbf{x}) \) depends on the input features, and the simpler mixture model case, where the weight \( \alpha \) is a constant.
\\\\
\textbf{Binary Classification}
\\\\
For binary classification, we define the dissimilarity measure \( d_{\text{pred}} \) in Equation~\eqref{eqn_MOE_obj} as the cross-entropy loss \( d_{\text{CE}} \), given by:
\begin{eqnarray}
d_{\text{CE}}(f_{\text{combined}}(\mathbf{x}), f_{\text{perf}}(\mathbf{x}))
&=&
- f_{\text{perf}}(\mathbf{x})
\log(f_{\text{combined}}(\mathbf{x}))
\nonumber \\
&&
- \left(1 - f_{\text{perf}}(\mathbf{x})\right)
\log\left(1 - f_{\text{combined}}(\mathbf{x})\right).
\end{eqnarray}

The fairness regularization term \( \mathcal{L}_{\text{fair}} \) is defined using the statistical parity gap and is given by:
\begin{equation}
\mathcal{L}_{\text{SP}}(f_{\text{combined}}) := \max_{\mathbf{a} \ne \mathbf{b}} \left| \mathbb{E}[f_{\text{combined}}(\mathbf{x}) \mid \mathbf{a}] - \mathbb{E}[f_{\text{combined}}(\mathbf{x}) \mid \mathbf{b}] \right|,
\label{eq:binary_fairness_metric}
\end{equation}

where \( \mathbf{a} \) and \( \mathbf{b} \) denote distinct groups of the sensitive attribute, and \( \mathbb{E}[f_{\text{combined}}(\mathbf{x}) \mid \mathbf{a}] \) is the average predicted probability of a positive outcome for group \( \mathbf{a} \). This term encourages demographic parity by minimizing the statistical parity gap across sensitive groups. Accordingly, the overall objective function becomes:

\begin{equation}
    \min_{\alpha, f_{\text{simp}}} 
    \left( \frac{1}{|\mathcal{D}_{pred}|} \sum_{\mathbf{x} \in \mathcal{D}_{pred}} 
    d_{\text{CE}}(f_{\text{combined}}(\mathbf{x}), f_{\text{perf}}(\mathbf{x})) 
    + \lambda \, \mathcal{L}_{\text{SP}}(f_{\text{combined}}; \mathcal{D}_{\text{sensitive}}) 
    \right),
\end{equation}

where \(\mathcal{L}_{\text{SP}}(f_{\text{combined}}; \mathcal{D}_{\text{sensitive}})\) is the empirical version of the penalty \(\mathcal{L}_{\text{SP}}(f_{\text{combined}})\) evaluated on \(\mathcal{D}_{\text{sensitive}}\).
\\
The convex-combination structure also preserves the validity of the predicted probabilities. Since
\(f_{\text{perf}}(\mathbf{x}), f_{\text{simp}}(\mathbf{x}) \in [0,1]\) and
\(\alpha(\mathbf{x}) \in [0,1]\), Equation~\eqref{eqn_weighted_model} implies
\[
f_{\text{combined}}(\mathbf{x}) \in [0,1].
\]
Thus, provided that both component models produce valid probability predictions, the combined model remains within the probability space without requiring an additional projection or clipping step.
\\\\
\noindent \textbf{Regression}
\\\\
For regression tasks, we define the dissimilarity measure \( d_{\text{pred}} \) as the squared error loss:
\begin{equation}
    d_{\text{SE}}(f_{\text{combined}}(\mathbf{x}), f_{\text{perf}}(\mathbf{x})) := 
    \left( f_{\text{combined}}(\mathbf{x}) - f_{\text{perf}}(\mathbf{x}) \right)^2.
\end{equation}
The fairness regularization term \( \mathcal{L}_{\text{fair}} \) is set to a variation of the Statistical Parity AUC penalty proposed by \cite{filippi2023local}, defined as:
\begin{eqnarray}
   && \mathcal{L}_{\text{SP\_AUC}}(f_{\text{combined}}) \nonumber \\
    &:=& \max_{\mathbf{a} \ne \mathbf{b}} 
    \int_0^1  \left| 
    \Pr(f_{\text{combined}}(\mathbf{x}) \geq Q(t) \mid \mathbf{a}) - 
    \Pr(f_{\text{combined}}(\mathbf{x}) \geq Q(t) \mid \mathbf{b}) 
    \right| \, dt
    \label{eq:regression_fairness_metric}
\end{eqnarray}

where \( \mathbf{a} \) and \( \mathbf{b} \) denote different groups of the sensitive attribute, \(t \in [0,1]\) is the quantile level, and \(Q(t)\) denotes the corresponding quantile of the prediction distribution. This measure evaluates the average disparity in predicted outcomes between groups across all possible thresholds. This fairness regularization term requires evaluating a one-dimensional integral, which can be efficiently computed using numerical methods. The resulting objective function is:
\begin{equation}
    \min_{\alpha, f_{\text{simp}}} 
    \left( \frac{1}{|D_{\mathrm{pred}}|} \sum_{\mathbf{x} \in D_{\mathrm{pred}}} 
    d_{\text{SE}}(f_{\text{combined}}(\mathbf{x}), f_{\text{perf}}(\mathbf{x})) 
    + \lambda \, \mathcal{L}_{\text{SP\_AUC}}(f_{\text{combined}}; \mathcal{D}_{\text{sensitive}}) 
    \right).
\end{equation}
For the global Mixture formulation, Equation~\eqref{eq:mixture_deviation} further gives
\begin{equation}
\label{eq:regression_mixture_deviation}
\left(
f_{\text{combined}}(\mathbf{x}) - f_{\text{perf}}(\mathbf{x})
\right)^2
=
(1-\alpha)^2
\left(
f_{\text{simp}}(\mathbf{x}) - f_{\text{perf}}(\mathbf{x})
\right)^2.
\end{equation}
Hence, under the squared-error fidelity measure, the deviation from the original predictor is explicitly controlled by the mixture weight and decreases quadratically as \(\alpha\) approaches one.
\\\\
\noindent \textbf{Survival Analysis}
\\\\
For survival analysis, let \( S_{\text{perf}}(t|\mathbf{x}) \) denote the survival function corresponding to feature $\mathbf{x}$ under the pre-trained model \( f_{\text{perf}} \), and \( S_{\text{combined}}(t|\mathbf{x}) \) denote the survival function under the combined model $f_{\text{combined}}$. 

An important structural property of the proposed formulation is that a convex combination of valid survival functions remains a valid survival function. In particular, the range constraint, monotonicity in time, and boundary conditions are preserved whenever the component models themselves produce valid survival functions. This is especially relevant in comparison with additive post-processing: a generic additive perturbation of a survival curve does not automatically preserve these properties and may therefore require additional constraints or transformations to ensure a valid survival output.

To quantify the dissimilarity between two survival functions over the time interval \([0, \tau]\), where \(\tau\) denotes the maximum follow-up time, we consider the integrated squared difference:
\begin{equation}
\int_0^\tau 
\left( S_{\text{combined}}(t|\mathbf{x}) - S_{\text{perf}}(t|\mathbf{x}) \right)^2 dt,
\end{equation}
which measures the cumulative discrepancy between the survival predictions of the combined and pre-trained models over time.

The fairness regularization term is defined via the group fairness penalty for survival functions \( \mathcal{L}_{\text{GF}} \) \cite{keya2021equitable, rahman2022fair, zhao2023fairness}. We first define group fairness at each time point \( t \) for the combined survival model \( S_{\text{combined}}(t|\mathbf{x}) \) as follows:

\begin{equation}
    \mathcal{L}_{\text{GF}}(S_{\text{combined}})(t) := 
    \max_{\mathbf{a} \in \mathcal{A}} \left| 
    \mathbb{E}_{\mathbf{x} \sim p_{\mathbf{a}}}[S_{\text{combined}}(t|\mathbf{x})] - 
    \mathbb{E}_{\mathbf{x} \sim p}[S_{\text{combined}}(t|\mathbf{x})]
    \right|
    \label{eq:survival_group_fairness_metric}
\end{equation}
where ${\cal A}$ is the set of sensitive attributes. 
Here, \( p_{\mathbf{a}}(\mathbf{x}) \) is the feature distribution for group \( \mathbf{a} \), and \( p(\mathbf{x}) \) is the population feature distribution, and
\begin{itemize}
    \item \( \mathbb{E}_{\mathbf{x} \sim p_{\mathbf{a}}}[S_{\text{combined}}(t|\mathbf{x})] := \int S_{\text{combined}}(t|\mathbf{x}) \, p_{\mathbf{a}}(\mathbf{x}) \, d\mathbf{x} \) is the expected survival probability at time \( t \) for group \( \mathbf{a} \),
    \item \( \mathbb{E}_{\mathbf{x} \sim p}[S_{\text{combined}}(t|\mathbf{x})] := \int S_{\text{combined}}(t|\mathbf{x}) \, p(\mathbf{x}) \, d\mathbf{x} \) is the expected survival probability at time \( t \) across the full population.
\end{itemize}
As a result, both expectations in Equation~\eqref{eq:survival_group_fairness_metric} can be approximated accordingly. 

Intuitively, \( \mathcal{L}_{\text{GF}}(S_{\text{combined}})(t) \) quantifies the extent to which each group’s average survival probability at time \( t \) differs from the overall population average. The greatest of these differences across all groups serves as a measure of unfairness at that specific time point. In other words, \( \mathcal{L}_{\text{GF}}(S_{\text{combined}})(t) \) captures the largest group-specific deviation from the population-level survival estimate at time \( t \).
To ensure fairness is accounted for across the entire time horizon, we incorporate a time-averaged fairness regularization by integrating 
\( \mathcal{L}_{\text{GF}}(S_{\text{combined}})(t) \) over time. This leads to the following objective function:
\begin{equation}
\label{eqn_opt_survival}
\min_{\alpha, f_{\text{simp}}} \left( 
\frac{1}{|D_{\mathrm{pred}}|} \sum_{\mathbf{x} \in D_{\mathrm{pred}}} \int_0^\tau 
\left( S_{\text{combined}}(t|\mathbf{x}) - S_{\text{perf}}(t|\mathbf{x}) \right)^2 dt 
+ \lambda \int_0^\tau \mathcal{L}_{\text{GF}}(S_{\text{combined}}; \mathcal{D}_{\text{sensitive}})(t) \, dt
\right).
\end{equation}
The evaluation of both integrals in Equation~\eqref{eqn_opt_survival} requires numerical integration. We employ a grid-based approach using evenly spaced points, where the number of grid points is adaptively increased until the integral estimates stabilize.
\\\\
\textbf{Optimization}
\\\\
For the classification setting, we employ the Limited-memory Broyden--Fletcher--Goldfarb--Shanno with Box constraints (L-BFGS-B) method \cite{zhu1997algorithm} for multivariate optimization of the gating function and/or the parameters of the simpler model \(f_{\text{simp}}\). Although the group-fairness objective may involve a maximum over group disparities, this reduces to the corresponding two-group disparity in the single-sensitive binary-group setting. For the regression setting, Powell's derivative-free optimization method \cite{powell1964efficient} is used for multivariate optimization, providing a suitable choice for the non-smooth SP-AUC fairness objective.

\section{An Ensemble Approach to Fairness, Two Pre-trained Models}
\label{sec_method_two_model}
This section explores a variation of the ensemble approach discussed previously, where two pre-trained models are available: one optimized for predictive performance, denoted by \( f_{\text{perf}} \), and another adjusted for fairness, denoted by \( f_{\text{fair}} \). The fairness-adjusted model may be obtained, for instance, by applying fairness-aware pre-processing to the input data prior to training, or by training a lower-complexity model on a restricted subset of non-sensitive features.

In this setting, both models are fixed, and the only learnable component is the weighting scheme—either a feature-dependent weight function in the MoE formulation or a constant weight in the simpler mixture model. The combined prediction model is then defined as:

\begin{itemize}
    \item \textbf{Mixture-of-Experts (MoE):}
    \begin{equation}
    f_{\text{combined}}(\mathbf{x}) = \alpha(\mathbf{x}) f_{\text{perf}}(\mathbf{x}) + (1 - \alpha(\mathbf{x})) f_{\text{fair}}(\mathbf{x}),
    \end{equation}
    where \( \alpha(\mathbf{x}) \) is a gating function that depends on the input features \( \mathbf{x} \).
    
    \item \textbf{Mixture Model:}
    \begin{equation}
    f_{\text{combined}}(\mathbf{x}) = \alpha f_{\text{perf}}(\mathbf{x}) + (1 - \alpha) f_{\text{fair}}(\mathbf{x}),
    \end{equation}
    where \( \alpha \in [0, 1] \) is a fixed scalar weight.
\end{itemize}
The task-dependent objective functions remain the same as those presented in Section \ref{sec_method_one_model}, with the key difference being that the optimization is now performed solely over the gating weight function \( \alpha(\mathbf{x}) \) in the MoE case, or over the constant mixture weight \( \alpha \) in the mixture model case.

\section{Experimental Setting and Results}

In this section, we describe the experimental setup, including model configurations, evaluation metrics, and datasets used for each task, along with the results obtained. 

% The implementation of our proposed methods is available at \url{https://github.com/noorazhaoz/fairness-ensemble-2025}.

\subsection{Datasets}
\label{sec:datasets}
We evaluate our framework on eight real-world datasets covering binary classification, regression, and survival analysis tasks. For binary classification, we use the \textit{COMPAS}~\cite{angwin2022machine}, \textit{Adult}~\cite{becker1996adult}, \textit{German Credit}~\cite{german_credit_data}, and \textit{Credit Card Default}~\cite{yeh2009comparisons} datasets. The \textit{COMPAS} and \textit{Adult} datasets are preprocessed following the methodology in~\cite{alghamdi2022beyond}, while the \textit{German Credit} dataset is processed using the AIF360 library~\cite{bellamy2018ai}. These datasets are used to predict recidivism risk, income level, credit risk, and credit card default, respectively. For regression, we use the \textit{Insurance} dataset~\cite{insurance_dataset}, where the prediction target is individual insurance charges based on demographic and health-related features, and the \textit{Law School GPA} dataset~\cite{wightman1998lsac}, where the target is standardized first-year law-school grade-point average based on academic and demographic information. For survival analysis, we use the \textit{WHAS} dataset~\cite{hosmer1999regression}, which models mortality among 481 patients following myocardial infarction, and the \textit{Employee Turnover} dataset~\cite{wijaya2020employee}, which models employment termination across 1,129 observations. The censoring rates are 48.23\% for \textit{WHAS} and 49.42\% for the \textit{Employee} dataset. We consider \texttt{gender} as the sensitive attribute across all datasets. For the \textit{COMPAS} and \textit{Adult} datasets, we additionally consider \texttt{race} to construct multi-sensitive-attribute settings. In \textit{COMPAS}, race is represented as Black or White, while in \textit{Adult}, it is represented as White or non-White. Continuous features are standardised prior to model training.

For each dataset, the input subset for the fairness-oriented model was selected from the available non-sensitive features following the minimum-cardinality procedure described in Section~\ref{sec:simpler_model_construction}. Candidate subsets were considered in increasing order of cardinality using the training data only. The minimum feature cardinality for which at least one candidate satisfied the fairness-admissibility criterion was first identified. When multiple subsets were admissible at this cardinality, predictive utility on the training data was used as a secondary selection criterion. The resulting subset was then fixed across all performance-model families, random seeds, combination mechanisms, and one- and two-pre-trained settings. Sensitive attributes were excluded from the fairness-oriented model, and neither validation nor test performance was used during feature selection.

The resulting restricted input space provides a deliberately low-complexity component with a different inductive bias from the performance-oriented model. Fairness is controlled through the explicit fairness regularization term and the learned combination mechanism, rather than by assuming that the selected features are inherently fair. Since non-sensitive features may still act as proxies for sensitive attributes such as gender or race, the selected feature subsets should not be interpreted as proxy-free. Table~\ref{tab:dataset} summarises the datasets, sensitive attributes, and feature subsets used in the fairness-oriented models.
\begin{table}[htbp]
    \centering
    \small
    \setlength{\tabcolsep}{1pt}
    \renewcommand{\arraystretch}{1.05}

    \caption{Summary of the datasets, sensitive attributes, and selected feature subsets used in the fairness-oriented models.}
    \label{tab:dataset}

    \begin{tabular}{clcccc}
        \toprule
        Task &
        Dataset &
        Size &
        \# Features &
        \makecell[c]{Sensitive\\Attribute(s)} &
        \makecell[c]{Selected Feature Subset\\for Fairness-oriented Model} \\
        \midrule

        \multirow{4}{*}{\makecell[c]{Binary\\Classification}}
        & \textit{COMPAS}
        & 7214
        & 6
        & \texttt{gender, race}
        & \texttt{[c\_charge\_degree]} \\

        & \textit{Adult}
        & 46447
        & 21
        & \texttt{gender, race}
        & \texttt{[education-num]} \\

        & \textit{German Credit}
        & 1000
        & 33
        & \texttt{gender}
        & \texttt{[Savings\_A62, Job\_A172]} \\

        & \textit{Credit Card Default}
        & 30000
        & 23
        & \texttt{gender}
        & \texttt{[PAY\_0]} \\

        \midrule

        \multirow{2}{*}{Regression}
        & \textit{Insurance}
        & 1338
        & 6
        & \texttt{gender}
        & \texttt{[smoker]} \\

        & \textit{Law School GPA}
        & 20798
        & 4
        & \texttt{gender}
        & \texttt{[ugpa]} \\

        \midrule

        \multirow{2}{*}{\makecell[c]{Survival\\Analysis}}
        & \textit{WHAS}
        & 481
        & 9
        & \texttt{gender}
        & \texttt{[lenstay]} \\

        & \textit{Employee}
        & 1129
        & 13
        & \texttt{gender}
        & \texttt{[extraversion]} \\

        \bottomrule
    \end{tabular}
\end{table}

\subsection{Tasks and Evaluation Metrics}
We evaluate our framework across three key machine learning tasks: binary classification, regression, and survival analysis. For \textbf{binary classification}, we use \texttt{accuracy} as the performance metric and assess fairness using \texttt{Demographic Parity (DP)} and \texttt{Equalized Odds (EO)}, both reported as gaps $\Delta$ (lower is better). \texttt{DP}$\Delta$ measures the range of positive prediction rates across sensitive groups~\cite{calders2009building}, while \texttt{EO}$\Delta$ measures the maximum groupwise disparity in true- and false-positive rates~\cite{hardt2016equality}. For \textbf{regression}, predictive performance is evaluated by \texttt{Mean Squared Error (MSE)}, and fairness is quantified using the \texttt{Statistical Parity AUC} defined in Equation~\eqref{eq:regression_fairness_metric}, which measures disparities in predicted values across sensitive groups over the full output range (lower is better). For \textbf{survival analysis}, we adopt \texttt{Harrell’s C-index}~\cite{harrell1982evaluating}, \texttt{Integrated Brier Score (IBS)}~\cite{graf1999assessment}, and \texttt{Area Under the Curve (AUC)}~\cite{uno2007evaluating} as performance metrics, and evaluate group fairness using the maximum groupwise deviation in expected survival probabilities, as defined in Equation~\eqref{eq:survival_group_fairness_metric} (lower is better).

\subsection{Model Configurations}
\label{sec_model_configurations}
For each task, we adopt a performance-optimized model as the base learner. Specifically, we use a Random Forest (RF) classifier~\cite{breiman2001random} or a Multi-Layer Perceptron (MLP) classifier~\cite{rumelhart1986learning} for binary classification, a Random Forest Regressor~\cite{segal2004machine} or an MLP regressor for regression, and a Random Survival Forest (RSF)~\cite{ishwaran2008random} for survival analysis. For binary classification, RF and MLP are selected to represent two distinct model families and inductive biases: a tree-based ensemble and a neural network. This allows us to examine how the proposed post-processing framework behaves across different model architectures rather than restricting the evaluation to a single type of predictor.

The RF-based models use 100 estimators, the MLP classifier has a single hidden layer with 20 units, and the RSF is configured with 1000 trees. For survival analysis, the performance-oriented survival functions are obtained directly from the fitted RSF, while the simpler fairness-oriented model uses a Cox proportional hazards model. All models are trained with fixed random seeds for reproducibility.

Our framework comprises two variants. The first variant (Section~\ref{sec_method_one_model}) uses a single pre-trained
performance-oriented model, while the second (Section~\ref{sec_method_two_model}) relies on two fixed, independently pre-trained models---one performance-oriented and one fairness-oriented. In both settings, the model outputs are combined using either a global mixture weight or an input-dependent MoE gating function. In the one-pre-trained setting, the simpler model and the combination mechanism are jointly optimized, whereas in the two-pre-trained setting, only the combination mechanism is learned while both component models remain fixed. The learnable components involve a relatively small number of parameters, particularly because the fairness-oriented model is constructed using a restricted input space.

% As described in Section~\ref{sec:datasets}, the fairness-oriented model is constructed using a single non-sensitive feature that is randomly selected before model training and then fixed across all experimental settings. This restricted input space provides a deliberately low-complexity component with a different inductive bias from the performance-oriented model. The feature is not selected to optimise fairness or predictive performance, and no assumption is made that it is intrinsically fair or free from proxy effects. The specific feature used for each dataset is listed in Table~\ref{tab:dataset}.

\subsection{Implementation and Baselines}
We compare four model configurations---one or two pre-trained models combined via either global mixture weighting or mixture-of-experts (MoE) weighting---as outlined in Section~\ref{sec_model_configurations}. 

For binary classification and regression tasks, we include \textsc{FRAPPÉ}~\cite{tifreafrappe} as a representative post-processing baseline. Additionally, we consider the Bayes-optimal fair classification approach with multiple sensitive features proposed by~\cite{yang2026bayes}, which we adapt to the fixed pre-trained prediction setting used in our framework.

We further compare against several established fairness baselines for binary classification. These include \textsc{Reject Option Classification (ROC)}~\cite{kamiran2012decision}, applied with \texttt{Demographic Parity} in our experiments, and \textsc{CalibratedEqOddsPostprocessing}~\cite{pleiss2017fairness}, which targets \texttt{Equalized Odds}. We also include two in-processing methods: \textsc{Reductions}~\cite{agarwal2018reductions}, which enforces fairness constraints through constrained optimization, and \textsc{HGR}~\cite{mary2019fairness}, which regularizes dependence between predictions and the sensitive attribute using the Hirschfeld--Gebelein--R\'enyi maximum correlation coefficient. In our experiments, HGR is instantiated with a demographic-parity-oriented regularization objective.

For survival analysis, we compare against the fairness-aware method in~\citep{zhao2023fairness}, which extends post-processing techniques to the time-to-event prediction setting.

Each dataset is partitioned into 80\% for training and 20\% for testing. Across all model configurations, we evaluate a predefined grid of the trade-off parameter \(\lambda\), which governs the balance between predictive performance and fairness. The corresponding model is optimized separately for each \(\lambda\) value, dataset, and experimental setting, allowing us to examine the resulting performance--fairness trade-off. 

\subsection{Core Results}

We present the main experimental results to evaluate the performance--fairness trade-offs achieved by the proposed post-processing framework. The analysis focuses on three aspects: fairness, predictive performance, and their joint trade-off. We first examine the fairness disparities obtained under different model configurations and then assess the corresponding changes in task-specific performance metrics, including accuracy, MSE, and C-index. Finally, we analyse the resulting Pareto frontiers across a range of \(\lambda\) values to characterise how the different methods navigate the performance--fairness trade-off.

\subsubsection{Binary Classification}

We evaluate our framework on three binary classification datasets, measuring predictive performance using \texttt{accuracy} and group fairness using the Demographic Parity (DP) and Equalized Odds (EO) gaps. All datasets are evaluated using both MLP and Random Forest as performance-oriented models under the single-sensitive-attribute (\texttt{gender}) setting. In addition, multi-sensitive-attribute (\texttt{gender + race}) settings are considered for the \textit{Adult} and \textit{COMPAS} datasets.

\begin{figure}[!htbp]
    \centering

    \includegraphics[width=0.98\textwidth]{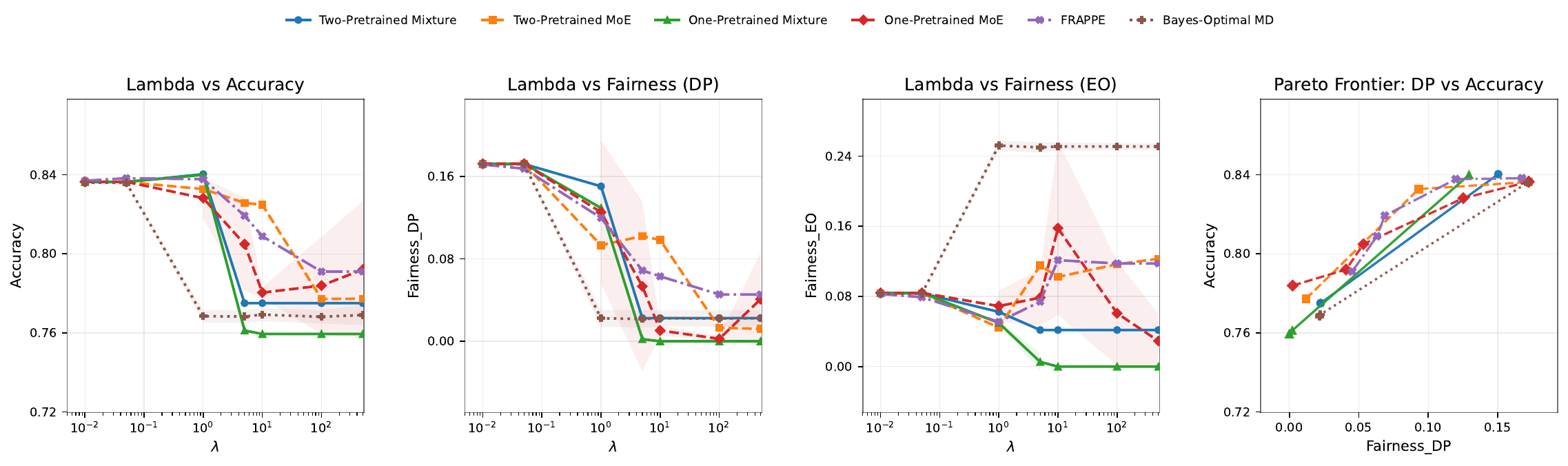}

    {RF -- Sensitive attribute: Gender}
    \vspace{0.6em}

    \includegraphics[width=0.98\textwidth]{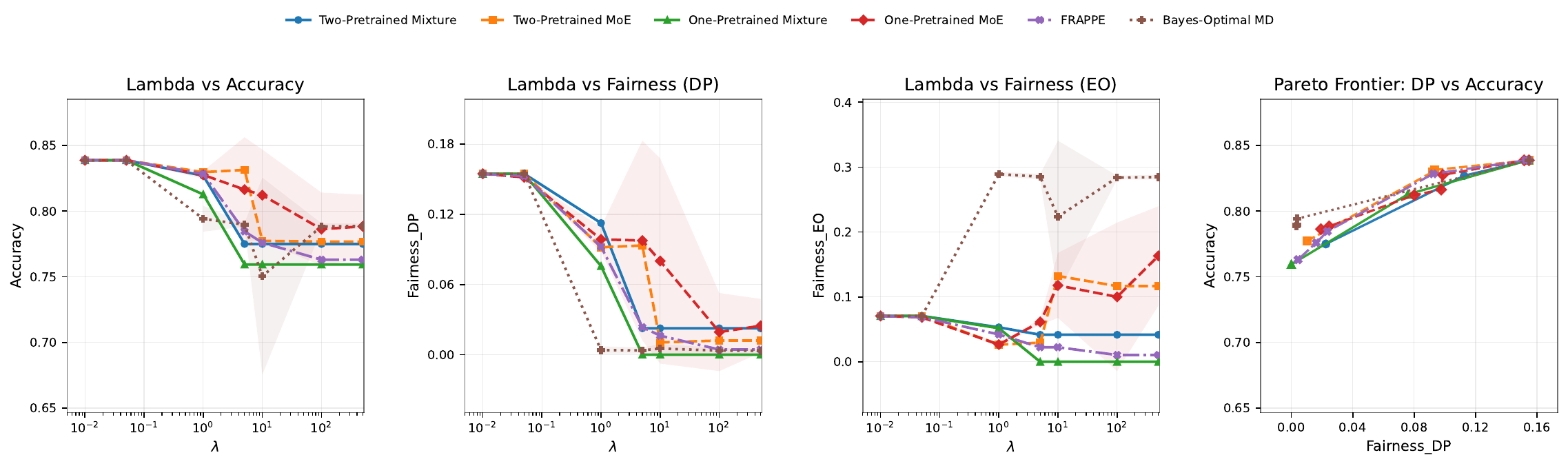}
    {MLP -- Sensitive attribute: Gender}
    \vspace{0.6em}

    \includegraphics[width=0.98\textwidth]{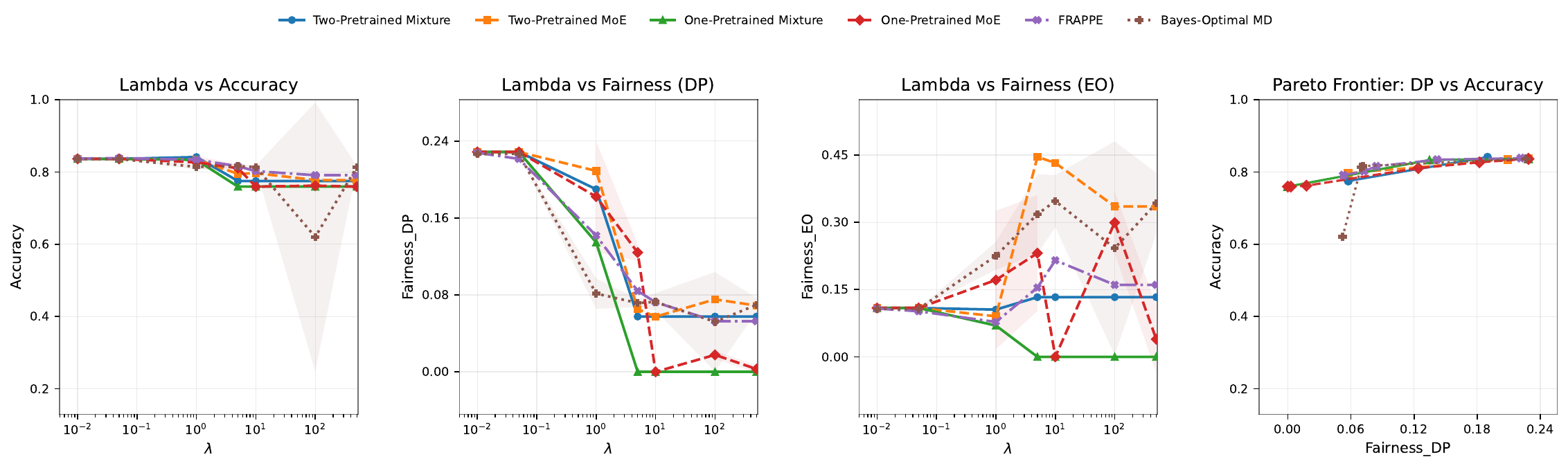}
    \par
    {RF -- Sensitive attributes: Gender and Race}
    \vspace{0.6em}

    \includegraphics[width=0.98\textwidth]{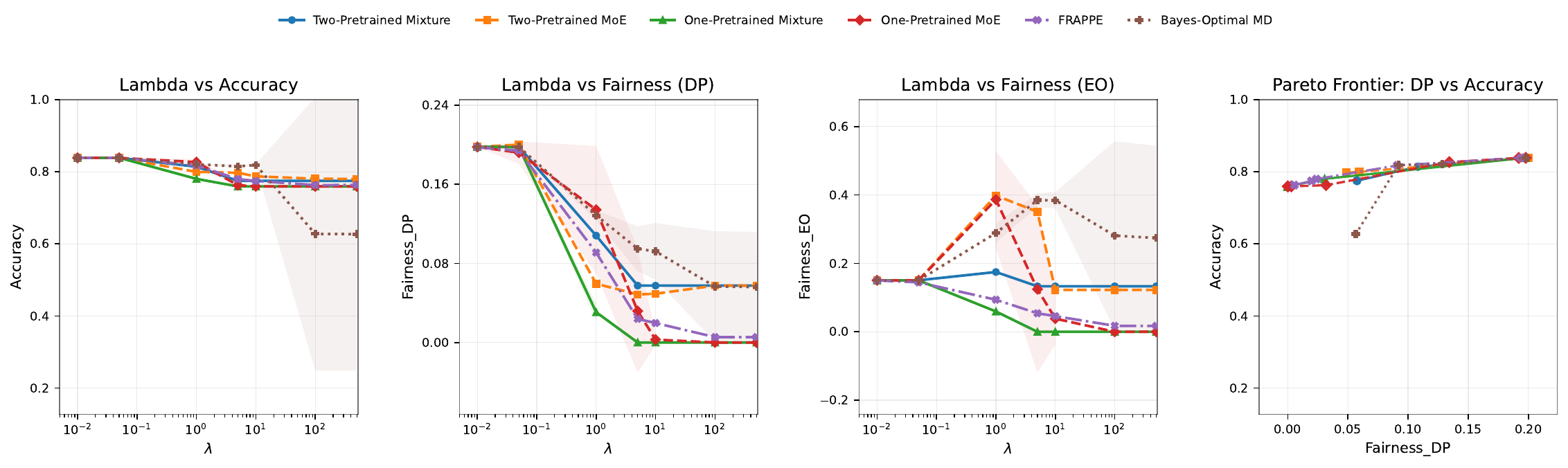}
    {MLP -- Sensitive attributes: Gender and Race}
    \caption{
    Performance--fairness trade-offs on the \textit{Adult} dataset across
    predictive backbones and sensitive-attribute settings. Each row presents the
    corresponding $\lambda$ sweeps and Pareto frontier, using
    $\lambda \in \{0.01, 0.05, 1, 5, 10, 100, 500\}$.
    }
    \label{fig:adult_tradeoffs}
\end{figure}

Figure~\ref{fig:adult_tradeoffs} summarizes the performance--fairness trade-offs obtained on the \textit{Adult} dataset across both predictive backbones and sensitive-attribute settings. Increasing \(\lambda\) produces a clear reduction in DP disparity for most proposed configurations, consistent with the fact that the regularization term is defined directly on the DP gap. The corresponding EO behavior is less monotonic, as EO is not explicitly optimized, although the proposed configurations generally maintain competitive EO values across the examined operating points. These fairness gains are accompanied by varying degrees of accuracy loss, depending on the predictive backbone and model configuration. The Pareto frontiers therefore indicate a continuum of feasible operating points rather than a uniformly optimal setting.

Relative to \textsc{FRAPPÉ}~\cite{tifreafrappe} and the mean-difference-based Bayes-optimal fair classification baseline (\textsc{Bayes-Optimal MD}) of~\cite{yang2026bayes}, the results likewise show no consistent dominance by a single method. The proposed configurations provide a broader set of intermediate trade-offs between predictive accuracy and DP disparity, with their relative advantage depending on the predictive backbone, sensitive-attribute definition, and chosen operating point.

\begin{figure}[!htbp]
    \centering

    \includegraphics[width=0.98\textwidth]{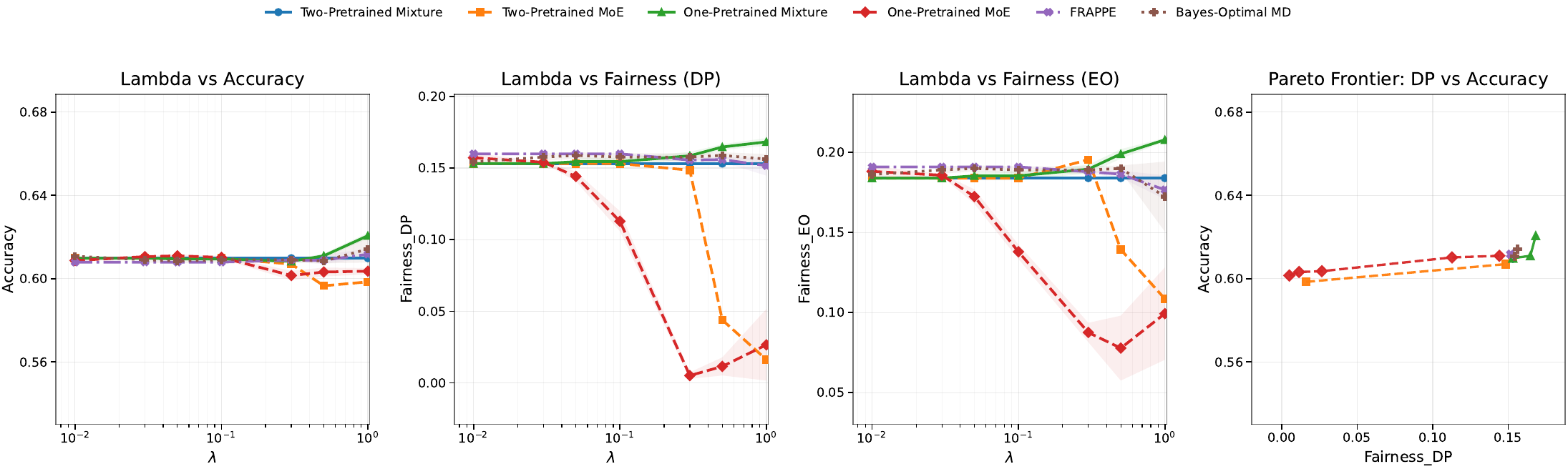}
    \par 
    {RF -- Sensitive attribute: Gender}
    \vspace{0.6em}

    \includegraphics[width=0.98\textwidth]{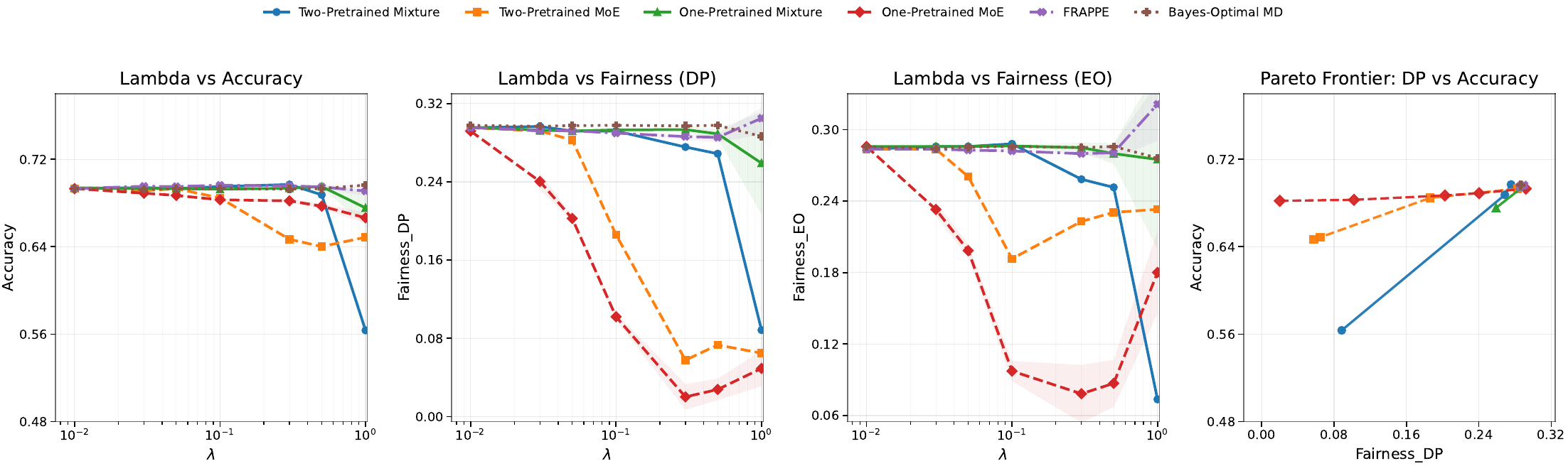}
    \par
    {MLP -- Sensitive attribute: Gender}
    \vspace{0.6em}

    \includegraphics[width=0.98\textwidth]{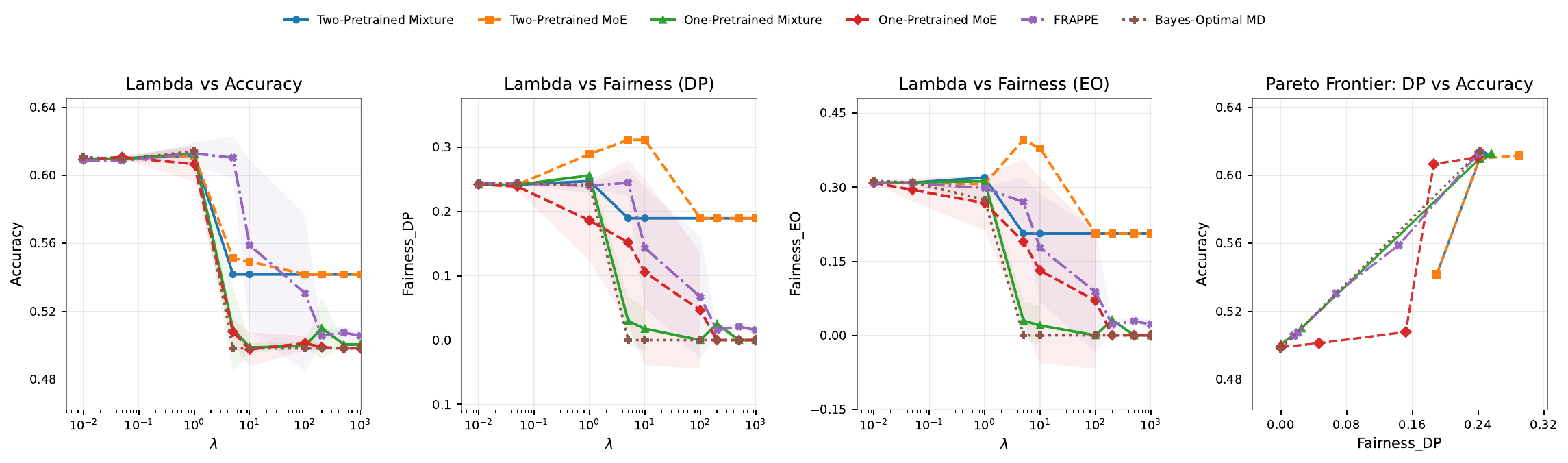}
    \par
    {RF -- Sensitive attributes: Gender and Race}
    \vspace{0.6em}

    \includegraphics[width=0.98\textwidth]{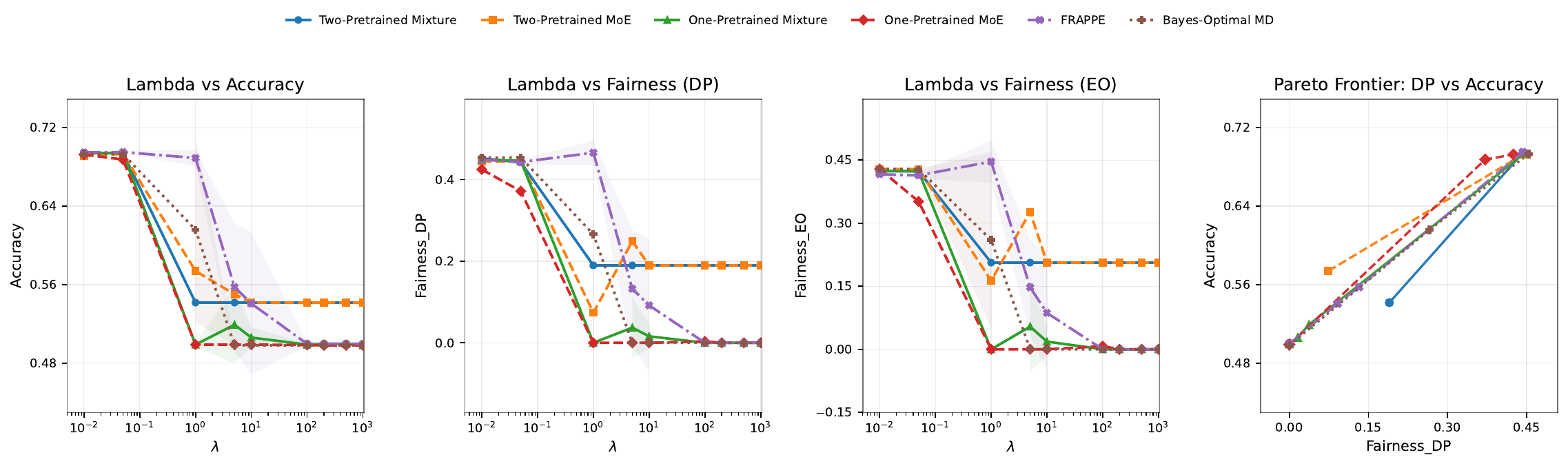}
    \par
    {MLP -- Sensitive attributes: Gender and Race}
    \caption{
    Performance--fairness trade-offs on the \textit{COMPAS} dataset across
    predictive backbones and sensitive-attribute settings. The four rows
    correspond to RF with gender, MLP with gender, RF with gender--race,
    and MLP with gender--race, respectively. Each row presents the
    corresponding $\lambda$ sweeps and Pareto frontier. For the
    single-sensitive setting (gender), we use
    $\lambda \in \{0.01, 0.03, 0.05, 0.1, 0.3, 0.5, 1.0\}$.
    For the multi-sensitive setting (gender--race), we use
    $\lambda \in \{0.01, 0.05, 1.0, 5.0, 10.0, 100.0, 200.0, 500.0, 1000.0\}$.
    }
    \label{fig:compas_tradeoffs}
\end{figure}

Figure~\ref{fig:compas_tradeoffs} summarizes the performance--fairness trade-offs on the \textit{COMPAS} dataset across both predictive backbones and sensitive-attribute settings. Compared with the \textit{Adult} results, the effect of increasing \(\lambda\) is more configuration-dependent. Several proposed configurations achieve substantial reductions in DP disparity, particularly at stronger regularization levels, while the corresponding EO changes are less systematic because EO is not directly optimized. Under the multi-sensitive-attribute setting, lower disparity can also be achieved, although this is often accompanied by a more pronounced reduction in predictive accuracy. The resulting Pareto frontiers therefore reflect a wider range of fairness--performance compromises across model configurations.

Compared with \textsc{FRAPPÉ} and \textsc{Bayes-Optimal MD}, the clearest advantage of the proposed framework appears in the single-sensitive-attribute setting. In particular, the One-Pretrained MoE configuration reduces the DP gap substantially under both RF and MLP while retaining accuracy close to the unregularized operating region, thereby reaching lower-disparity points than \textsc{FRAPPÉ} with a relatively limited performance penalty. Under the multi-sensitive-attribute setting, some proposed configurations also extend into lower-DP regions than \textsc{FRAPPÉ}, although the associated accuracy loss becomes more pronounced and the advantage over \textsc{Bayes-Optimal MD} is less consistent.

\begin{figure*}[h]
    \centering

    \includegraphics[width=0.95\textwidth]{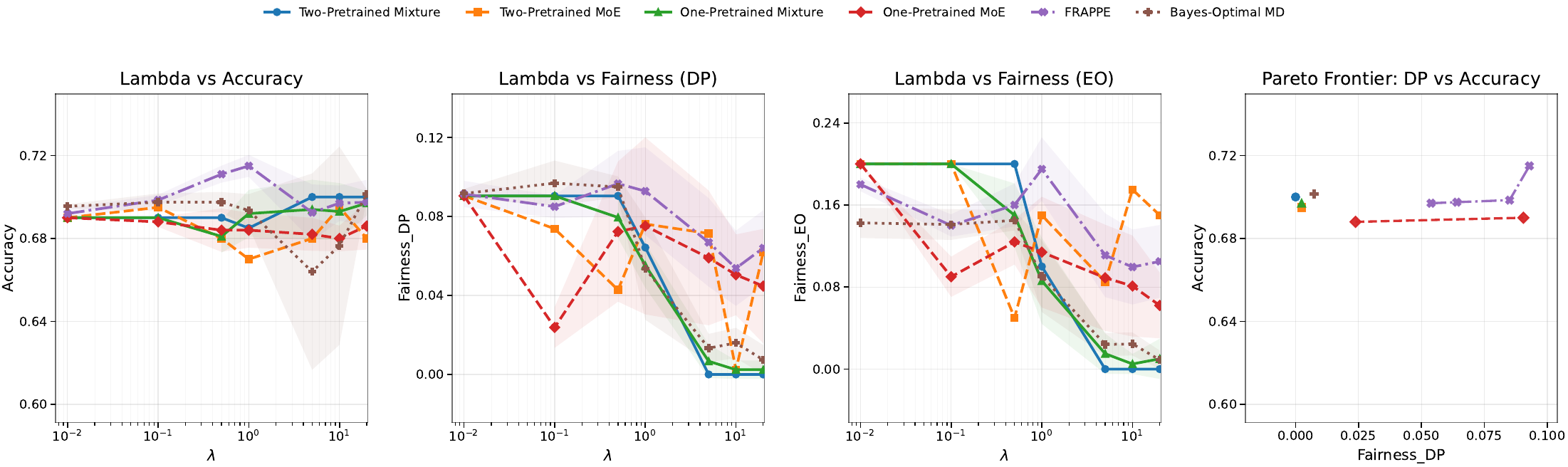}
    \par
    {RF -- Sensitive attribute: Gender}
    \vspace{0.6em}

    \includegraphics[width=0.95\textwidth]{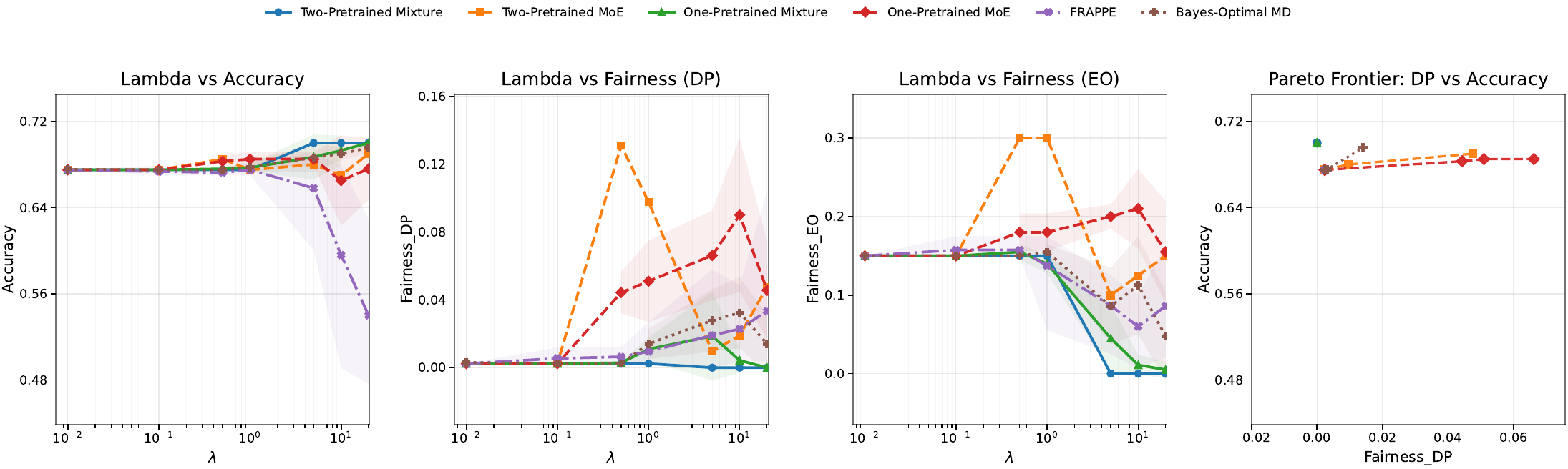}
    \par
    {MLP -- Sensitive attribute: Gender}
    \caption{
    Performance--fairness trade-offs on the \textit{German} dataset using
    gender as the sensitive attribute. The two rows correspond to RF and MLP
    predictive backbones, respectively. Each row presents the corresponding
    $\lambda$ sweeps and Pareto frontier, using
    $\lambda \in \{0.01, 0.1, 0.5, 1, 5, 10, 20\}$.
    }
    \label{fig:german_tradeoffs}
\end{figure*}

The \textit{German Credit} dataset serves as a compact yet fairness-sensitive benchmark, using \texttt{gender} as the sensitive attribute (Figure~\ref{fig:german_tradeoffs}). Given the small test set of only 200 instances, the reported fairness metrics are relatively sensitive to changes in a limited number of subgroup-specific predictions, which contributes to the less smooth and occasionally non-monotonic behavior across \(\lambda\). This sensitivity is particularly visible for EO, which can vary even when accuracy and DP remain close to the unregularized operating region. Nevertheless, several proposed configurations achieve meaningful reductions in DP disparity while preserving competitive predictive accuracy, especially under the RF backbone. Under MLP, the trade-offs are more variable, with stronger regularization sometimes producing larger changes in both fairness and accuracy. Relative to \textsc{FRAPPÉ}, the proposed configurations provide additional low- and intermediate-disparity operating points, although the preferred operating region is less stable than on the larger datasets.

\begin{figure*}[h]
    \centering

    \includegraphics[width=0.95\textwidth]{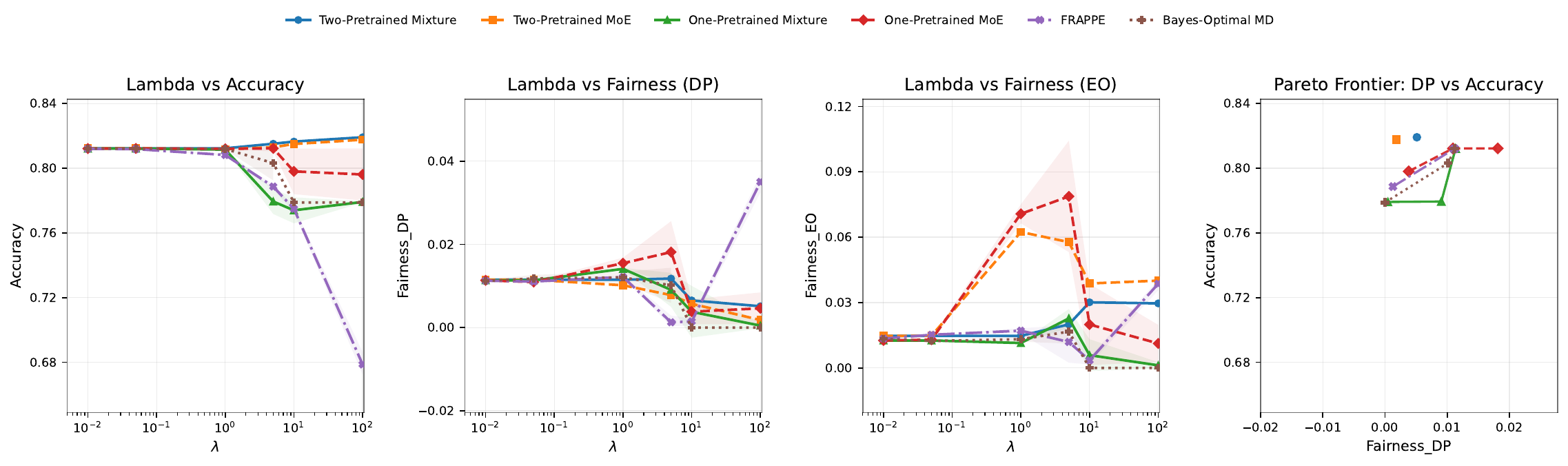}
    \par
    {RF -- Sensitive attribute: Gender}
    \vspace{0.6em}

    \includegraphics[width=0.95\textwidth]{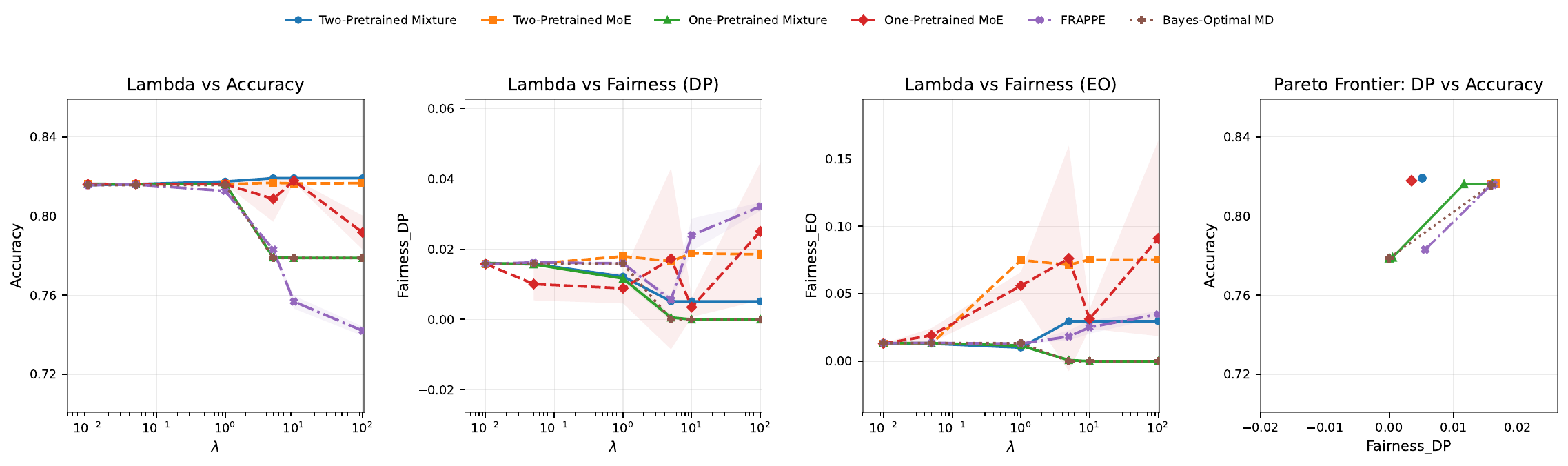}
    \par
    {MLP -- Sensitive attribute: Gender}
    \caption{
    Performance--fairness trade-offs on the \textit{Credit} dataset using
    gender as the sensitive attribute. The two rows correspond to RF and MLP
    predictive backbones, respectively. Each row presents the corresponding
    $\lambda$ sweeps and Pareto frontier, using
    $\lambda \in \{0.01, 0.05, 1, 5, 10, 100\}$.
    }
    \label{fig:credit_tradeoffs}
\end{figure*}

Figure~\ref{fig:credit_tradeoffs} reports the corresponding results on the \textit{Credit} dataset. The initial DP disparity is already small, and several proposed configurations maintain or slightly reduce this gap while preserving accuracy close to the unregularized level, particularly under moderate regularization. The corresponding EO behaviour is more configuration-dependent, with some settings remaining close to their initial values while others deteriorate under stronger regularization. Under both RF and MLP, the main performance loss occurs only for a subset of strongly regularized configurations.

Relative to \textsc{FRAPPÉ}, the proposed methods achieve more favourable trade-offs on this dataset: \textsc{FRAPPÉ} provides little additional reduction in DP disparity while incurring a more noticeable loss in accuracy, whereas several proposed configurations preserve near-baseline predictive performance at similarly low or lower DP levels.

\paragraph{Comparison with Additional Baselines.}
Table~\ref{tab:bin_four_ds_dp_eo} provides additional comparisons with established fairness-aware baselines under the single-sensitive-attribute setting. Because the proposed framework is evaluated over a predefined discrete grid of \(\lambda\) values, we do not tune or interpolate \(\lambda\) to match the fairness level of each baseline. Instead, we examine whether any evaluated operating point Pareto-dominates a baseline, namely, whether it achieves no lower predictive accuracy and no larger DP$\Delta$ or EO$\Delta$, with at least one strict improvement.

The proposed framework yields Pareto-dominating operating points in several dataset--backbone settings. On \textit{Adult}, such points are obtained relative to both \textsc{Reject Option Classification} and \textsc{CalibratedEqOddsPostprocessing} under RF and MLP, while the very small DP gaps achieved by \textsc{Reductions} are not simultaneously matched in predictive accuracy and EO. On \textit{COMPAS}, the comparison is more challenging: no evaluated RF configuration dominates the additional baselines, whereas under MLP the proposed framework dominates \textsc{CalibratedEqOddsPostprocessing} and HGR (EO-targeted). On \textit{German}, several baselines are dominated, particularly under MLP, although the small test set makes the empirical fairness gaps more sensitive to individual predictions. On \textit{Credit}, the proposed framework dominates all three RF baselines and both post-processing baselines under MLP, while \textsc{Reductions} and HGR remain non-dominated at their respective operating points.

These results do not indicate uniform superiority over all competing fairness interventions. Rather, they show that the predefined set of proposed configurations contains competitive, and in several cases Pareto-dominating, solutions without baseline-specific tuning of \(\lambda\). Baselines that are not dominated represent alternative fairness--performance compromises rather than necessarily outperforming the proposed framework on all three reported criteria. At the same time, varying \(\lambda\) provides multiple operating points from the same performance-oriented base model, allowing the resulting trade-off between predictive utility and group disparity to be examined directly.

Finally, \textsc{CalibratedEqOddsPostprocessing} does not provide explicit control over intersectional fairness, while \textsc{Reject Option Classification} is formulated for binary sensitive attributes, limiting their direct extension to the multi-sensitive settings considered here \cite{kearns2018preventing,mehrabi2021survey}. The proposed framework can instead incorporate multiple sensitive attributes through the corresponding group definition, as evaluated separately on \textit{Adult} and \textit{COMPAS}.

\newcommand{\perf}{$\uparrow$}
\newcommand{\fair}{$\downarrow$}
\newcommand{\na}{—}

\begin{table}[htbp]
    \centering
    \footnotesize
    \setlength{\tabcolsep}{3.8pt}
    \caption{Binary classification results across four datasets and two base models
    (RF/MLP). Predictive performance is measured by accuracy ($\uparrow$), while
    fairness is reported using demographic parity difference (DP$\Delta$) and
    equalized odds difference (EO$\Delta$), where lower values indicate better
    fairness ($\downarrow$). The final column indicates whether at least one
    operating point of the proposed framework, evaluated over the predefined
    discrete grid of $\lambda$ values, Pareto-dominates the corresponding
    baseline, i.e., achieves no lower predictive accuracy and no larger
    DP$\Delta$ or EO$\Delta$, with at least one strict improvement. No
    interpolation or baseline-specific tuning of $\lambda$ is performed.}
    \label{tab:bin_four_ds_dp_eo}

    \begin{tabular}{lllcccc}
        \toprule
        \textbf{Dataset} & \textbf{Base} & \textbf{Method}
        & \textbf{Perf}~(\perf)
        & \textbf{DP$\Delta$}~(\fair)
        & \textbf{EO$\Delta$}~(\fair)
        & \textbf{Ours dom.} \\
        \midrule

        %================ Adult ================
        \multirow{10}{*}{Adult}
        & \multirow{5}{*}{RF}
          & ROC                            & 0.7752 & 0.0468 & 0.1329 & \checkmark \\
        & & CalibratedEqOddsPostprocessing & 0.8327 & 0.2575 & 0.6033 & \checkmark \\
        & & Reductions                     & 0.8404 & 0.0119 & 0.3163 & -- \\
        & & HGR (DP-targeted)              & \na    & \na    & \na    & \na \\
        & & HGR (EO-targeted)              & \na    & \na    & \na    & \na \\

        \cmidrule(lr){2-7}

        & \multirow{5}{*}{MLP}
          & ROC                            & 0.7835 & 0.0409 & 0.1923 & \checkmark \\
        & & CalibratedEqOddsPostprocessing & 0.8316 & 0.2572 & 0.6001 & \checkmark \\
        & & Reductions                     & 0.8392 & 0.0187 & 0.3037 & -- \\
        & & HGR (DP-targeted)              & 0.8448 & 0.1446 & 0.0600 & -- \\
        & & HGR (EO-targeted)              & 0.8423 & 0.1334 & 0.0536 & -- \\
        \midrule

        %================ COMPAS ================
        \multirow{10}{*}{COMPAS}
        & \multirow{5}{*}{RF}
          & ROC                            & 0.6268 & 0.0438 & 0.0311 & -- \\
        & & CalibratedEqOddsPostprocessing & 0.6477 & 0.2200 & 0.1815 & -- \\
        & & Reductions                     & 0.6439 & 0.0033 & 0.0542 & -- \\
        & & HGR (DP-targeted)              & \na    & \na    & \na    & \na \\
        & & HGR (EO-targeted)              & \na    & \na    & \na    & \na \\

        \cmidrule(lr){2-7}

        & \multirow{5}{*}{MLP}
          & ROC                            & 0.6600 & 0.0338 & 0.0387 & -- \\
        & & CalibratedEqOddsPostprocessing & 0.6255 & 0.1609 & 0.1571 & \checkmark \\
        & & Reductions                     & 0.6610 & 0.0405 & 0.0195 & -- \\
        & & HGR (DP-targeted)              & 0.5824 & 0.0234 & 0.0453 & -- \\
        & & HGR (EO-targeted)              & 0.5909 & 0.1025 & 0.0976 & \checkmark \\
        \midrule

        %================ German ================
        \multirow{10}{*}{German}
        & \multirow{5}{*}{RF}
          & ROC                            & 0.6530 & 0.0355 & 0.0674 & \checkmark \\
        & & CalibratedEqOddsPostprocessing & 0.6550 & 0.1129 & 0.0833 & \checkmark \\
        & & Reductions                     & 0.7350 & 0.0288 & 0.0430 & -- \\
        & & HGR (DP-targeted)              & \na    & \na    & \na    & \na \\
        & & HGR (EO-targeted)              & \na    & \na    & \na    & \na \\

        \cmidrule(lr){2-7}

        & \multirow{5}{*}{MLP}
          & ROC                            & 0.6100 & 0.0290 & 0.0579 & \checkmark \\
        & & CalibratedEqOddsPostprocessing & 0.6800 & 0.0832 & 0.0579 & \checkmark \\
        & & Reductions                     & 0.6550 & 0.0489 & 0.0874 & \checkmark \\
        & & HGR (DP-targeted)              & 0.6500 & 0.0000 & 0.0000 & \checkmark \\
        & & HGR (EO-targeted)              & 0.6500 & 0.0000 & 0.0000 & \checkmark \\
        \midrule

        %================ Credit ================
        \multirow{10}{*}{Credit}
        & \multirow{5}{*}{RF}
          & ROC                            & 0.7598 & 0.0040 & 0.0262 & \checkmark \\
        & & CalibratedEqOddsPostprocessing & 0.8045 & 0.0557 & 0.1063 & \checkmark \\
        & & Reductions                     & 0.8090 & 0.0114 & 0.0645 & \checkmark \\
        & & HGR (DP-targeted)              & \na    & \na    & \na    & \na \\
        & & HGR (EO-targeted)              & \na    & \na    & \na    & \na \\

        \cmidrule(lr){2-7}

        & \multirow{5}{*}{MLP}
          & ROC                            & 0.7670 & 0.0210 & 0.0200 & \checkmark \\
        & & CalibratedEqOddsPostprocessing & 0.8087 & 0.0591 & 0.1238 & \checkmark \\
        & & Reductions                     & 0.8133 & 0.0010 & 0.0410 & -- \\
        & & HGR (DP-targeted)              & 0.8168 & 0.0077 & 0.0237 & -- \\
        & & HGR (EO-targeted)              & 0.8168 & 0.0040 & 0.0208 & -- \\
        \bottomrule
    \end{tabular}

    \vspace{0.3em}
    {\footnotesize
    \emph{Note.} HGR requires differentiable logits and is therefore implemented
    with MLP only; RF entries are reported as n/a. A checkmark indicates Pareto
    dominance by at least one evaluated operating point of the proposed framework.}
\end{table}

\subsubsection{Regression}

\begin{figure*}[h]
    \centering

    \includegraphics[width=0.95\textwidth]{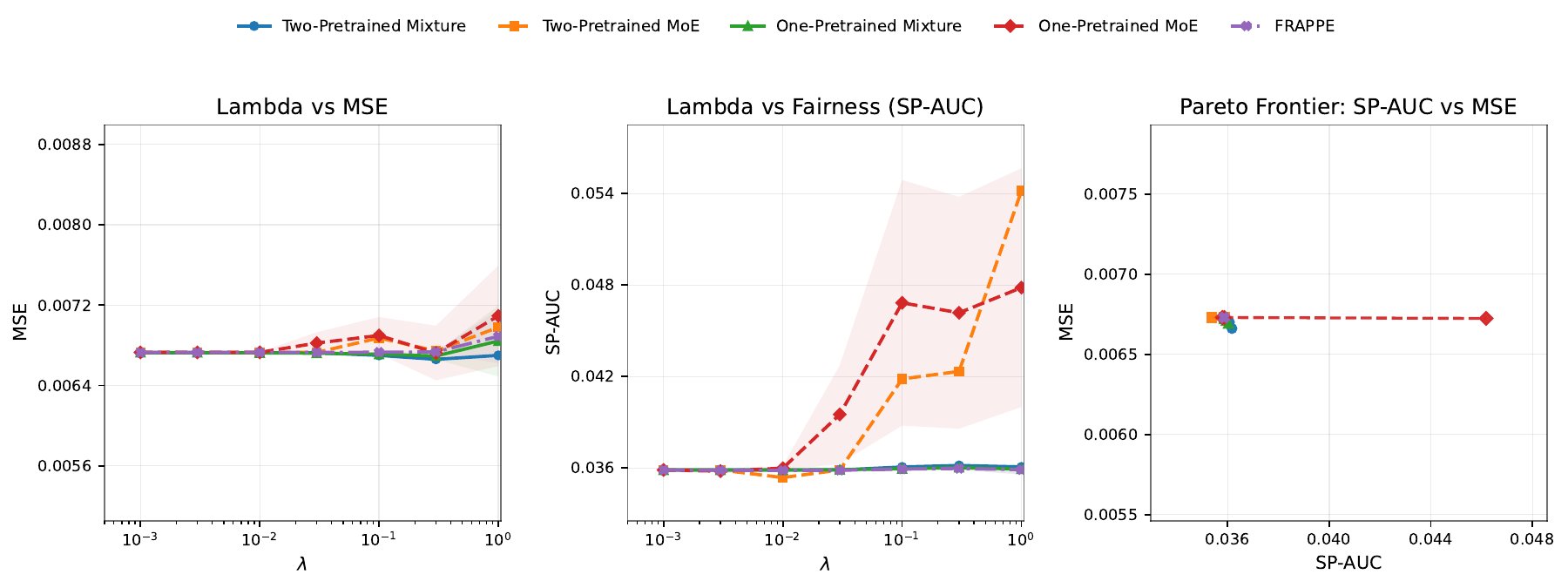}
    \par
    {RF -- Sensitive attribute: Gender}
    \vspace{0.6em}
    
    \includegraphics[width=0.95\textwidth]{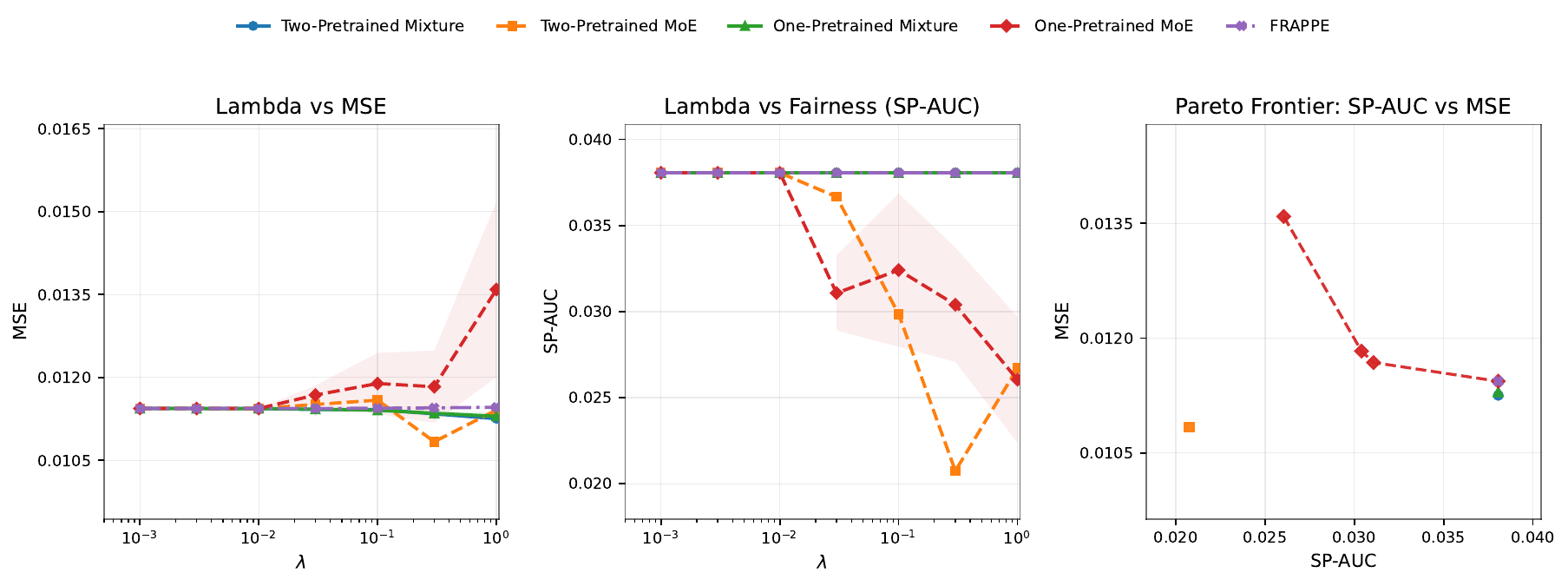}
    \par
    {MLP -- Sensitive attribute: Gender}
    
    \caption{
    Performance--fairness trade-offs on the \textit{Insurance} dataset. The panels show the variation in MSE and SP-AUC across the tested values of the regularization parameter, together with the resulting Pareto frontier. We evaluate
    $\lambda \in \{0.001, 0.003, 0.01, 0.03, 0.1, 0.3, 1\}$.
    }
    \label{fig:insurance_tradeoffs}
\end{figure*}

\begin{figure*}[h]
    \centering

    \includegraphics[width=0.95\textwidth]{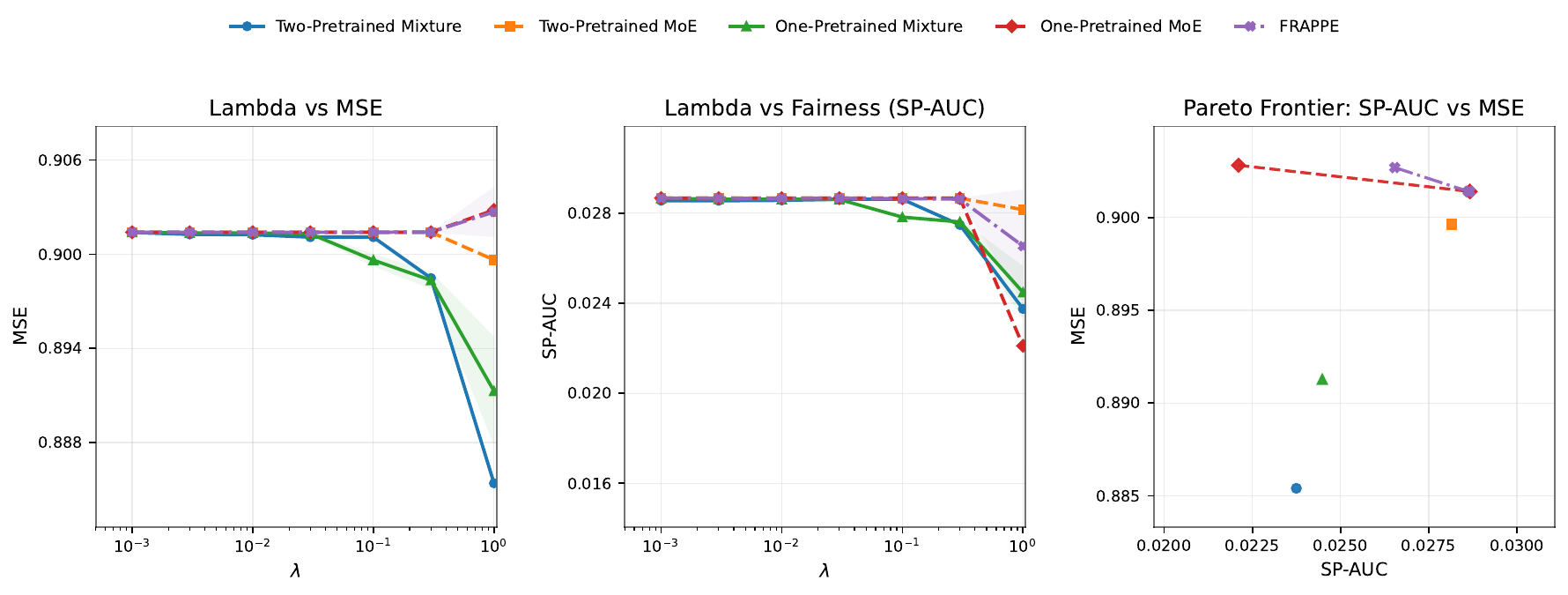}
    \par
    {RF -- Sensitive attribute: Gender}
    \vspace{0.6em}
    
    \includegraphics[width=0.95\textwidth]{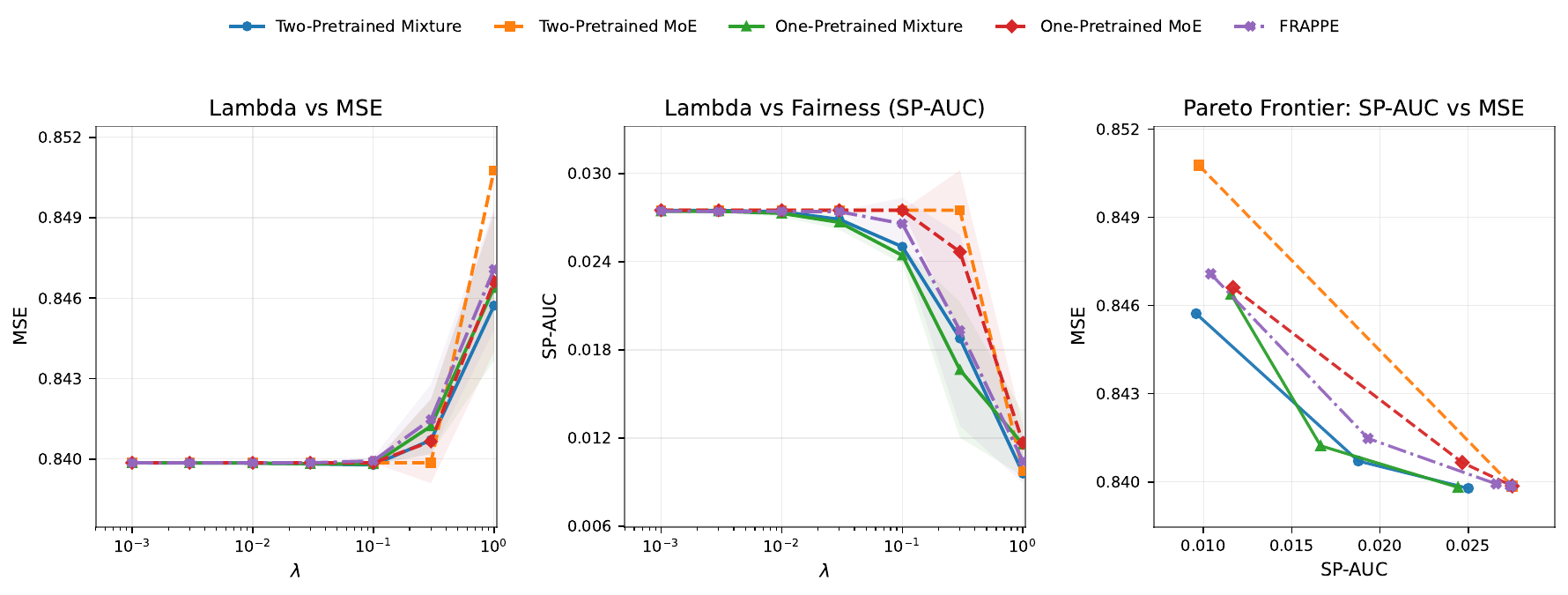}
    \par
    {MLP -- Sensitive attribute: Gender}
    
    \caption{
    Performance--fairness trade-offs on the \textit{Law School GPA} dataset. The panels show the variation in MSE and SP-AUC across the tested values of the regularization parameter, together with the resulting Pareto frontier. We evaluate
    $\lambda \in \{0.001, 0.003, 0.01, 0.03, 0.1, 0.3, 1, 3, 10, 30\}$.
    }
    \label{fig:lawschool_tradeoffs}
\end{figure*}

Figures~\ref{fig:insurance_tradeoffs} and~\ref{fig:lawschool_tradeoffs} present the regression results on the \textit{Insurance} and \textit{Law School GPA} datasets, respectively. On \textit{Insurance}, \textsc{FRAPPÉ} shows little change in SP-AUC under either backbone. The proposed Mixture variants exhibit similarly limited variation, whereas the MoE variants show a stronger dependence on the predictive backbone: they achieve clear SP-AUC reductions with MLP, but do not yield corresponding improvements with RF. The results on \textit{Law School GPA} are more consistently favourable across both backbones. Under RF, increasing \(\lambda\) leads to a clear reduction in SP-AUC, with several proposed configurations moving from approximately \(0.027\) to around \(0.01\) while incurring only a modest increase in MSE. The proposed mixture-based variants also provide a broader set of low-disparity operating points than \textsc{FRAPPÉ}, which remains concentrated in a narrower region of the trade-off space. Under MLP, several proposed configurations achieve lower SP-AUC while maintaining comparable predictive error, and the two mixture-based variants in particular move toward regions with both lower disparity and lower MSE at stronger regularization levels. Overall, the regression results indicate that the effectiveness of the proposed configurations is dataset- and backbone-dependent, while still providing useful fairness--performance trade-offs beyond those obtained by \textsc{FRAPPÉ}.

\subsubsection{Survival Analysis}

\begin{figure}[t]
    \centering
    \includegraphics[width=0.95\textwidth]{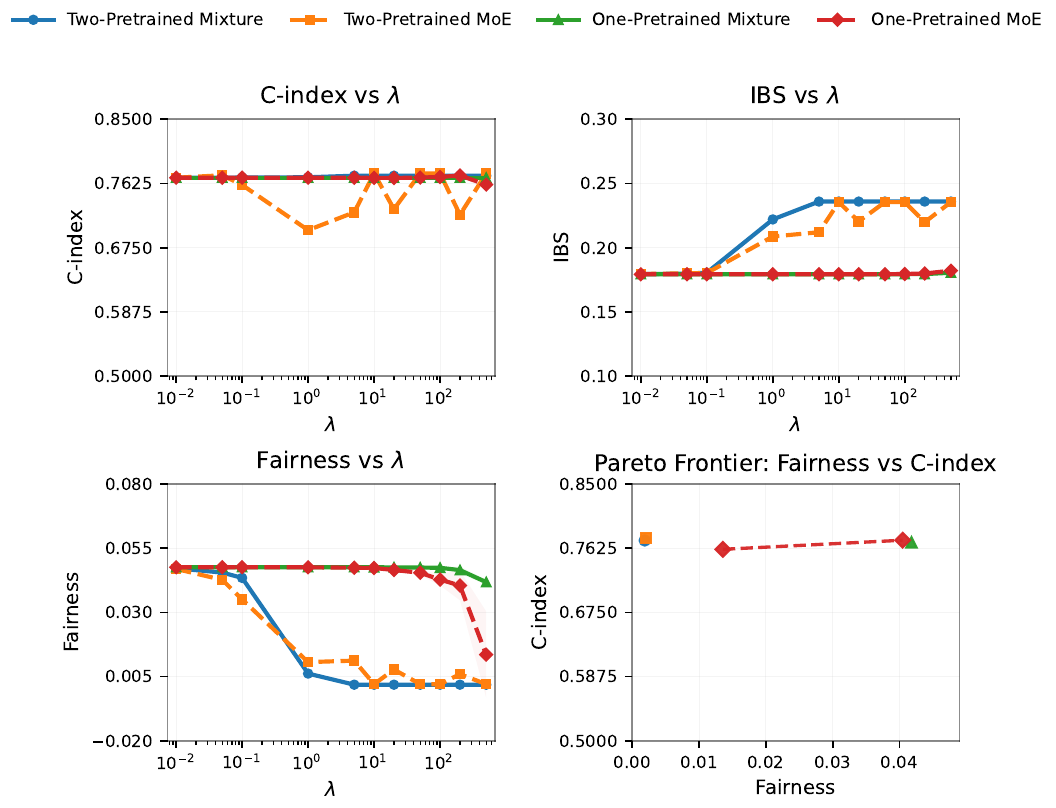}
    \caption{\footnotesize
    Performance--fairness trade-offs on the \textit{WHAS} survival dataset
    using a Random Survival Forest (RSF) as the performance-oriented model.
    We evaluate
    $\lambda \in \{0.01,0.05,0.1,1,5,10,20,50,100,200,500\}$.
    The reference result from Zhao and Ng~\cite{zhao2023fairness} is
    C-index $=0.7703$, IBS $=0.1783$, AUC $=0.8644$, and fairness $=0.036$.}
    \label{fig:whas_4}
\end{figure}

\begin{figure}[t]
    \centering
    \includegraphics[width=0.95\textwidth]{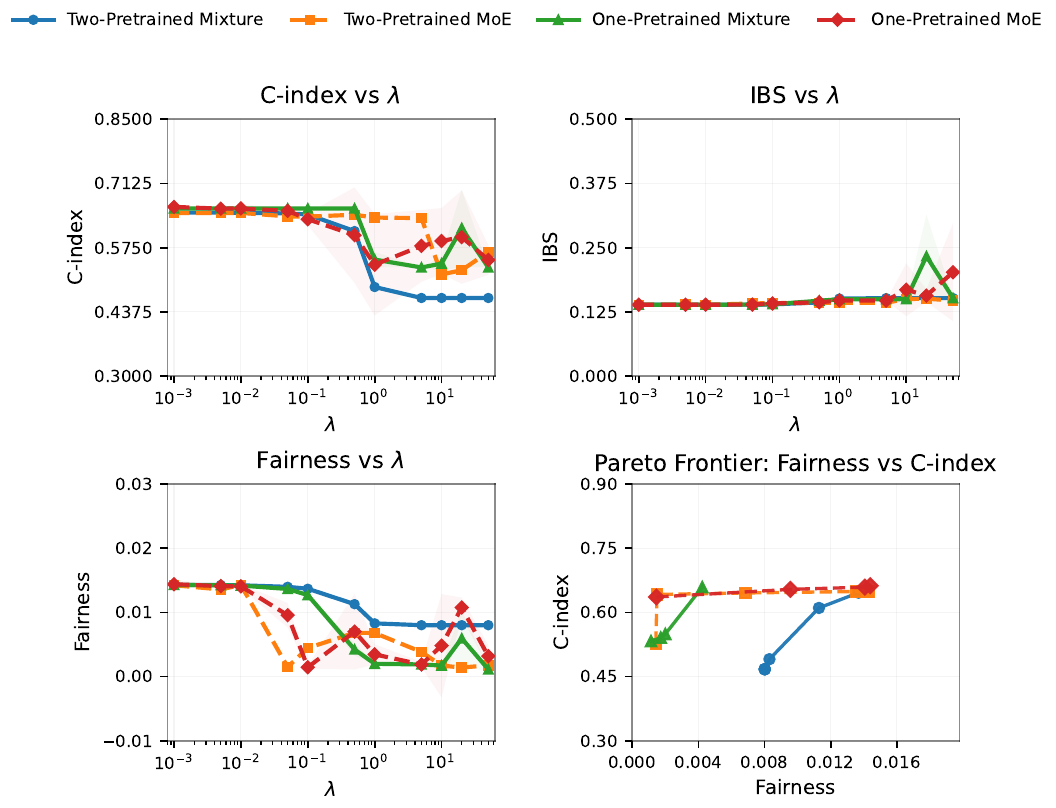}
    \caption{\footnotesize
    Performance--fairness trade-offs on the \textit{Employee} survival dataset
    using a Random Survival Forest (RSF) as the performance-oriented model.
    We evaluate
    $\lambda \in \{0.001,0.005,0.01,0.05,0.1,0.5,1,5,10,20,50\}$.
    The reference result from Zhao and Ng~\cite{zhao2023fairness} is
    C-index $=0.6602$, IBS $=0.1783$, AUC $=0.7423$, and fairness $=0.0088$.}
    \label{fig:employee_4}
\end{figure}

We evaluate survival prediction on the \textit{WHAS} and \textit{Employee}
datasets (Figures~\ref{fig:whas_4} and Figures~\ref{fig:employee_4}) using \texttt{C-index}, \texttt{AUC}, and \texttt{IBS} for predictive performance, and group fairness disparity for fairness assessment. On \textit{WHAS}, the \texttt{1-pretrained} configurations maintain relatively stable predictive performance, while the \texttt{2-pretrained} methods achieve larger fairness reductions at the cost of greater performance loss. On \textit{Employee}, the four configurations start from similar performance levels, with several methods reducing disparity as $\lambda$ increases while maintaining comparable predictive performance over part of the regularisation range. Overall, the results show that $\lambda$ provides a tunable fairness--performance trade-off, with the resulting behaviour depending on both the dataset and model configuration. Several configurations also achieve lower fairness disparity than the reference results of Zhao and Ng~\cite{zhao2023fairness}.

\subsection{Robustness to Deeper MLP Architectures}

To assess robustness to increased model capacity, we further evaluate representative
one-pretrained variants of the proposed framework using deeper three-layer
\texttt{MLP} backbones on the \textit{Adult} and \textit{COMPAS} datasets
(Table~\ref{tab:mlp_depth_robustness}). For \textit{Adult}, we report the
One-Pretrained Mixture, while for \textit{COMPAS} we report the
One-Pretrained MoE. 

The results show that the proposed framework continues to produce fairness--performance trade-offs under the deeper architecture. On Adult, the One-Pretrained Mixture attains lower DP and EO than \textsc{FRAPPÉ} while retaining comparable predictive accuracy at the selected operating point. On COMPAS, the One-Pretrained MoE attains lower DP and slightly lower EO while also achieving higher accuracy than the corresponding \textsc{FRAPPÉ} result.

\begin{table}[t]
\centering
\caption{
Robustness to increased MLP depth on the Adult and COMPAS datasets.
Results are reported as the bootstrap mean $\pm$ the half-width of the 95\% confidence interval.
}
\label{tab:mlp_depth_robustness}

\scriptsize
\setlength{\tabcolsep}{2pt}

\begin{tabular}{llcccc}
\toprule
Dataset & Method & $\lambda$ & Accuracy & DP & EO \\
\midrule

Adult
& One-Pretrained Mixture
& 1
& $\mathbf{0.8244 \pm 0.0034}$
& $\mathbf{0.0855 \pm 0.0032}$
& $\mathbf{0.0457 \pm 0.0057}$ \\

Adult
& \textsc{FRAPPÉ}
& 1
& $0.8295 \pm 0.0003$
& $0.0987 \pm 0.0005$
& $0.0682 \pm 0.0020$ \\

\midrule

COMPAS
& One-Pretrained MoE
& 1
& $\mathbf{0.6499 \pm 0.0103}$
& $\mathbf{0.0191 \pm 0.0164}$
& $\mathbf{0.1315 \pm 0.0384}$ \\

COMPAS
& \textsc{FRAPPÉ}
& 5
& $0.6016 \pm 0.0452$
& $0.1049 \pm 0.0758$
& $0.1352 \pm 0.0620$ \\

\bottomrule
\end{tabular}

\end{table}

\subsection{t-SNE Visualization of Instance-Adaptive Weight Distributions (MoE)}

\begin{figure}[htbp]
    \centering
    \subfigure[Single-sensitive setting (\texttt{gender})]{
        \includegraphics[width=0.48\textwidth]{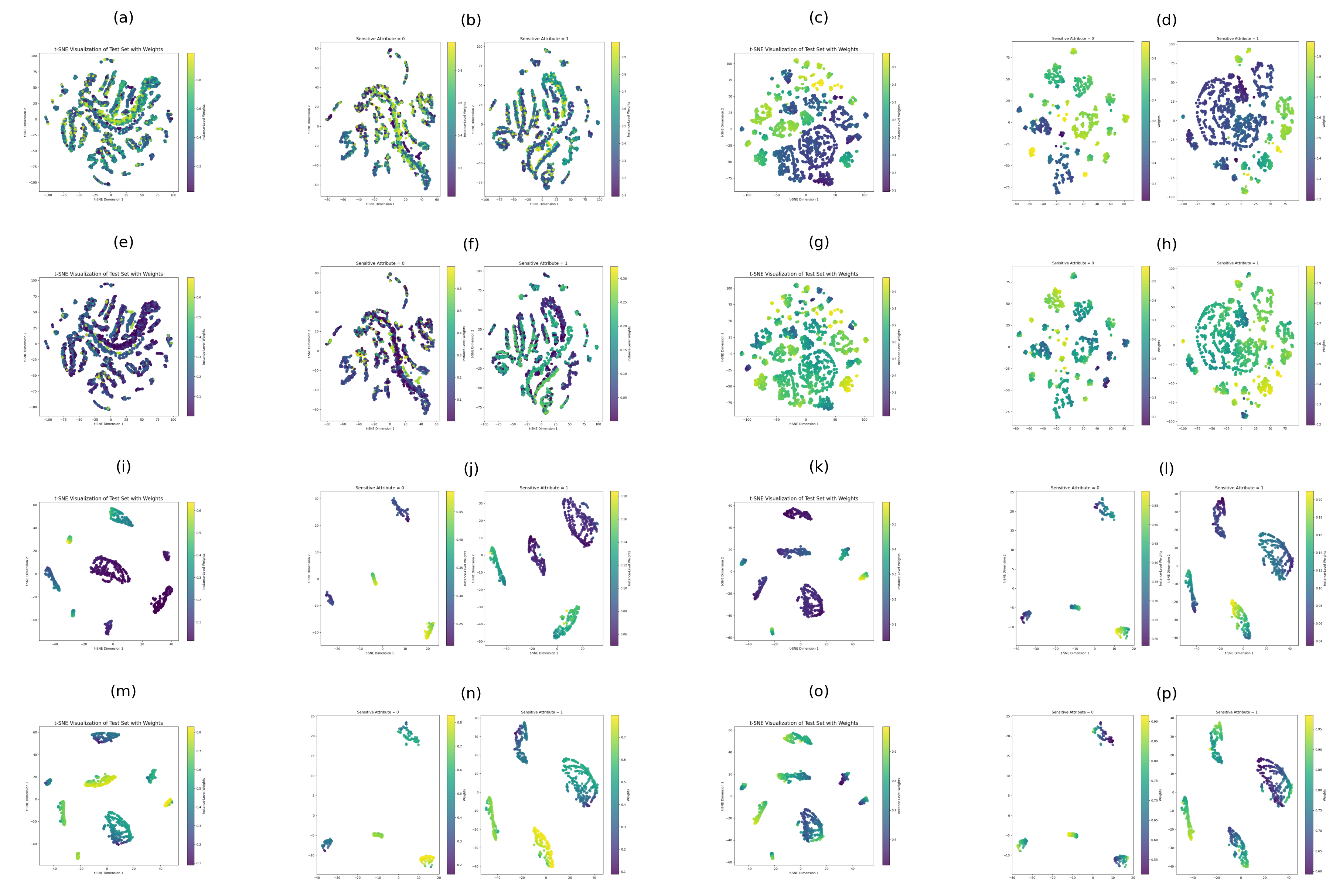}
        \label{fig:tsne_single}
    }
    \hfill
    \subfigure[Multi-sensitive setting (\texttt{gender + race})]{
        \includegraphics[width=0.46\textwidth]{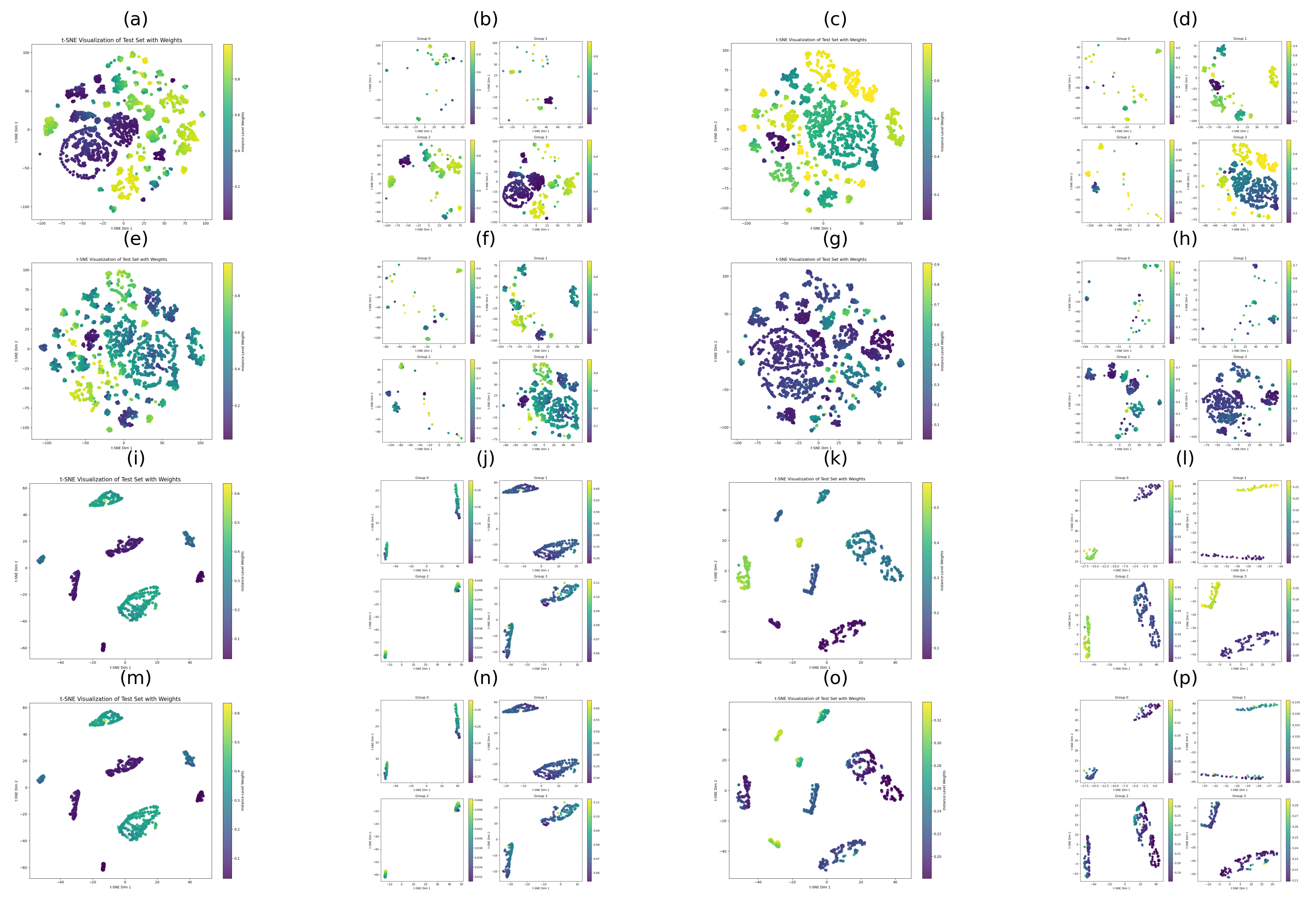}
        \label{fig:tsne_multisen}
    }
    \caption{\footnotesize
    t-SNE visualizations of instance-level weights on the \textit{Adult} and \textit{COMPAS} datasets under both single- and multi-sensitive attribute settings. In each subfigure, the first two rows ((a)–(h)) correspond to \textit{Adult}, and the last two ((i)–(p)) to \textit{COMPAS}. For each dataset, the first row uses Random Forest (RF) as the performance model, and the second uses a 1-layer Multi-Layer Perceptron (MLP). Each pair of columns shows overall (left) and subgroup-based (right) visualizations of instance-level weights. The left half of each row corresponds to the \textit{1-pretrained MoE} model, and the right half to the \textit{2-pretrained MoE} model.
    }
    \label{fig:tsne_adult_compas}
\end{figure}

\begin{figure}[htbp]
\centering
\resizebox{0.9\textwidth}{!}{%
\begin{minipage}{\textwidth}
\centering
\subfigure[GERMAN: One/Two Models — Overall, By \texttt{gender}]{
  \includegraphics[width=0.21\textwidth]{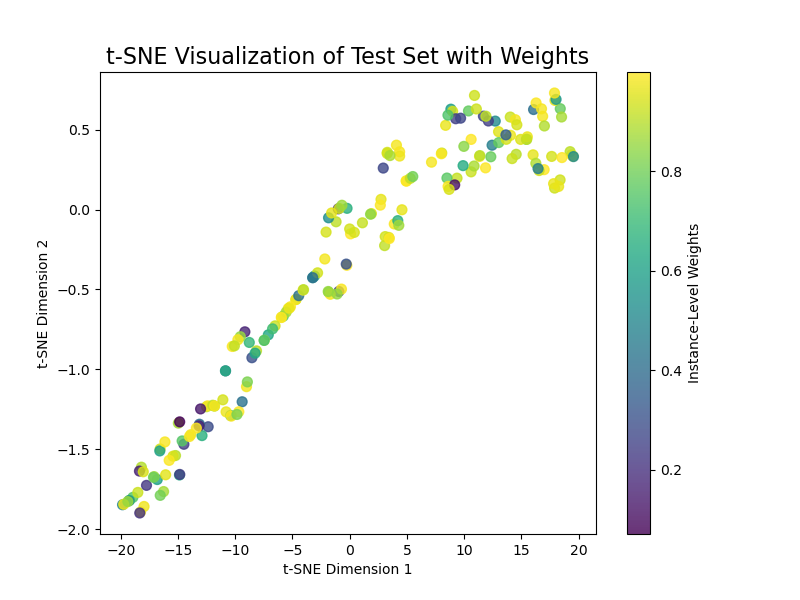}
  \includegraphics[width=0.31\textwidth]{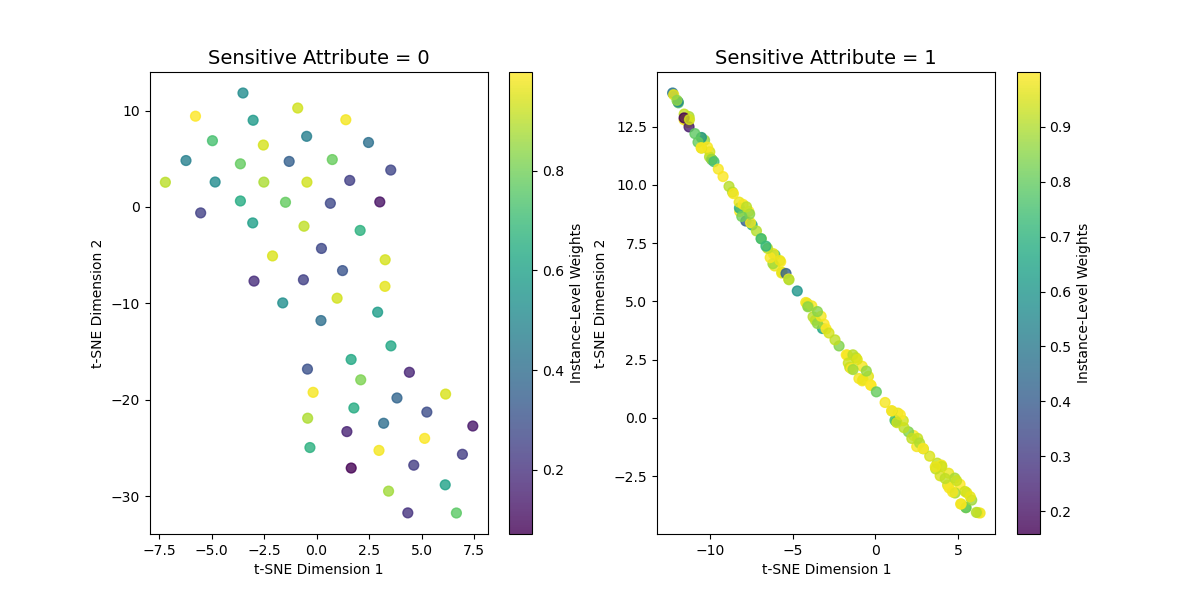}
  \includegraphics[width=0.20\textwidth]{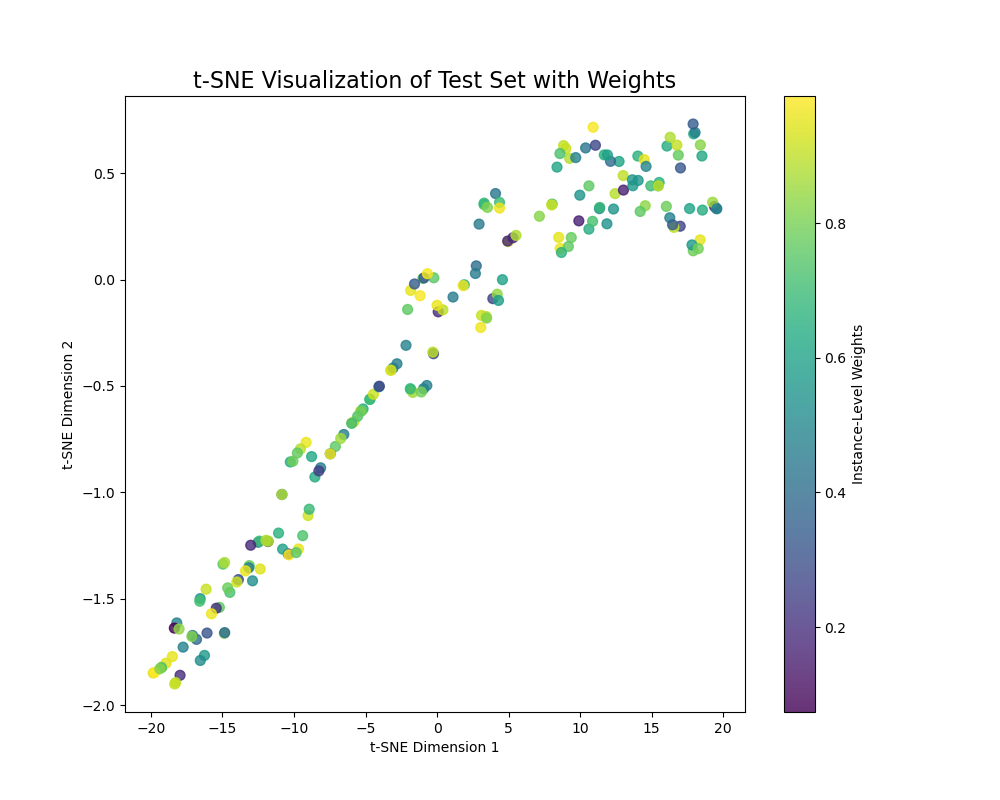}
  \includegraphics[width=0.29\textwidth]{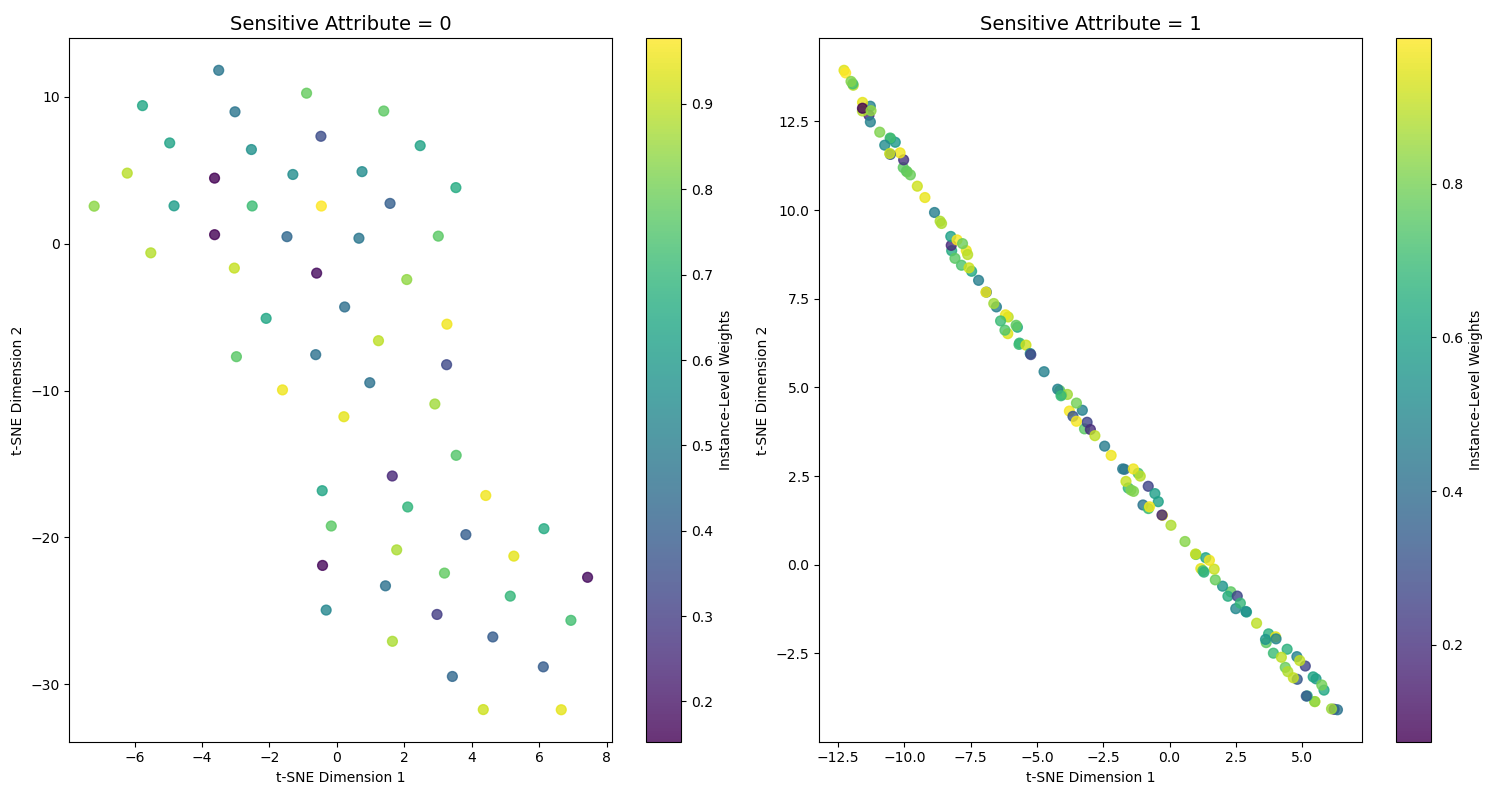}
}
\subfigure[Insurance: One/Two Models — Overall, By \texttt{gender}]{
  \includegraphics[width=0.21\textwidth]{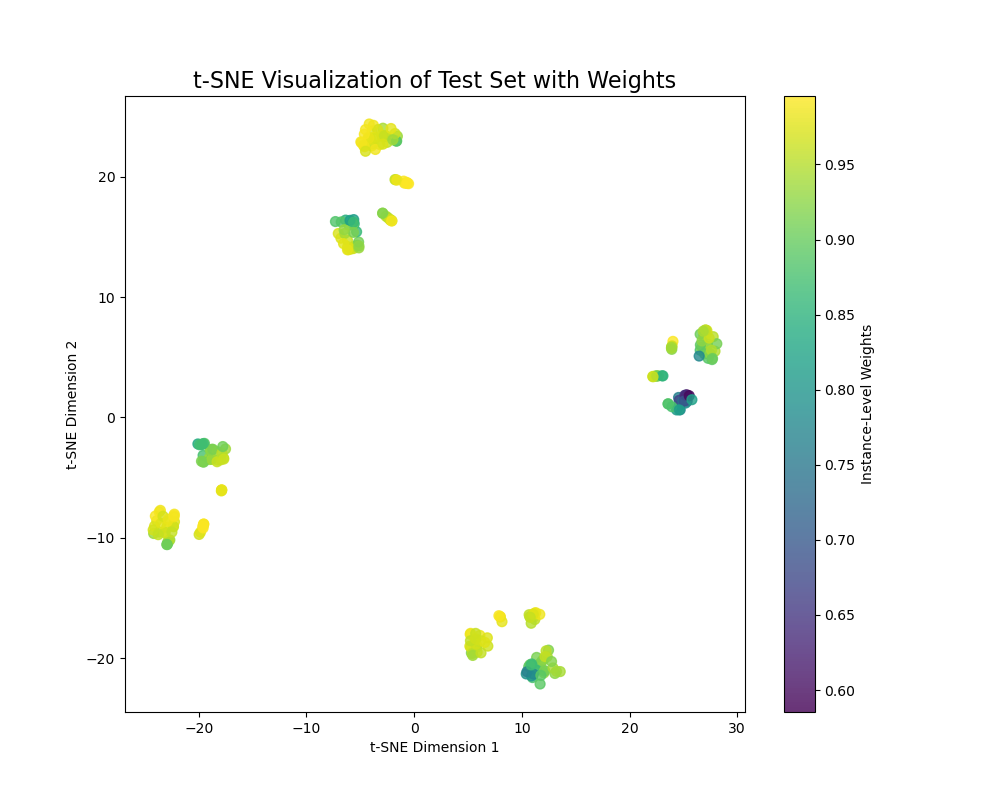}
  \includegraphics[width=0.29\textwidth]{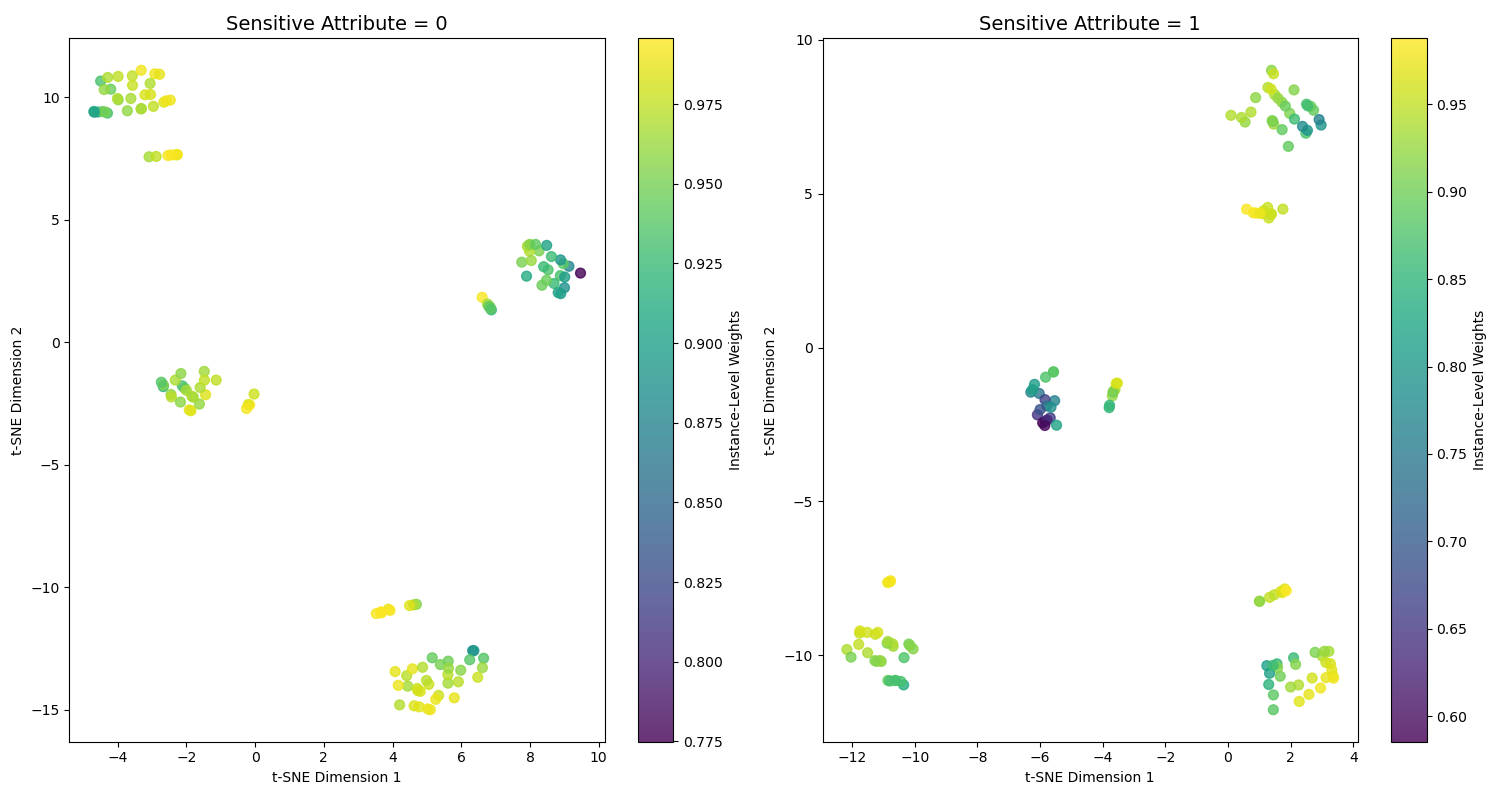}
  \includegraphics[width=0.21\textwidth]{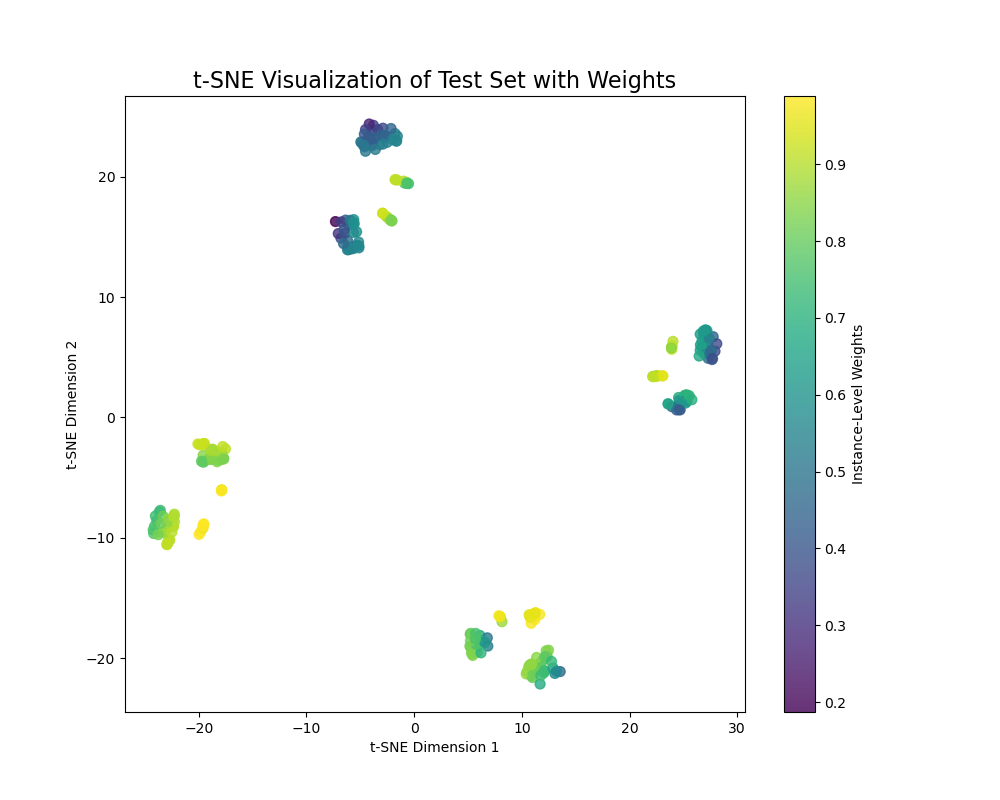}
  \includegraphics[width=0.29\textwidth]{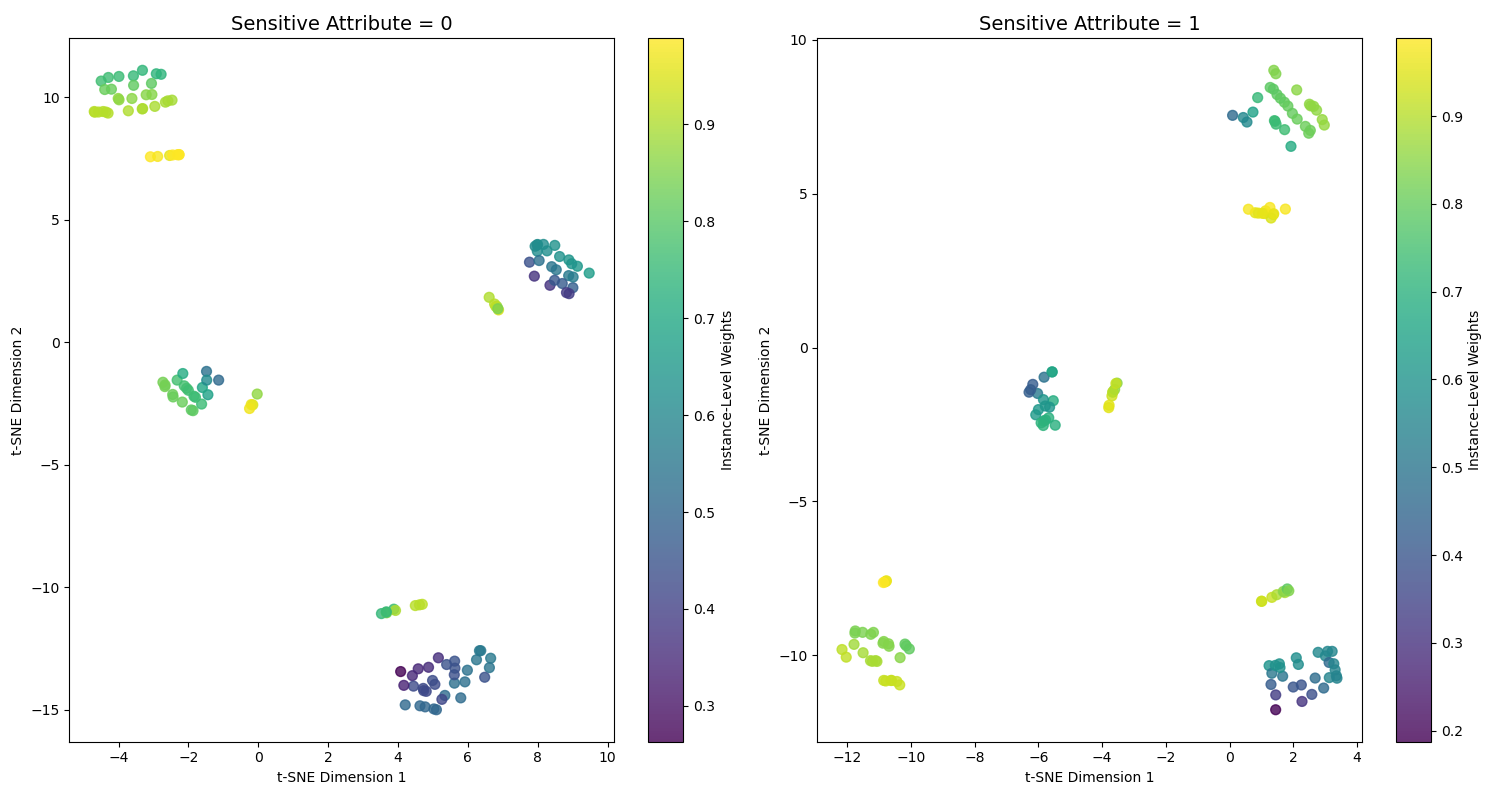}
}
\subfigure[WHAS: One/Two Models — Overall, By \texttt{gender}]{
  \includegraphics[width=0.18\textwidth]{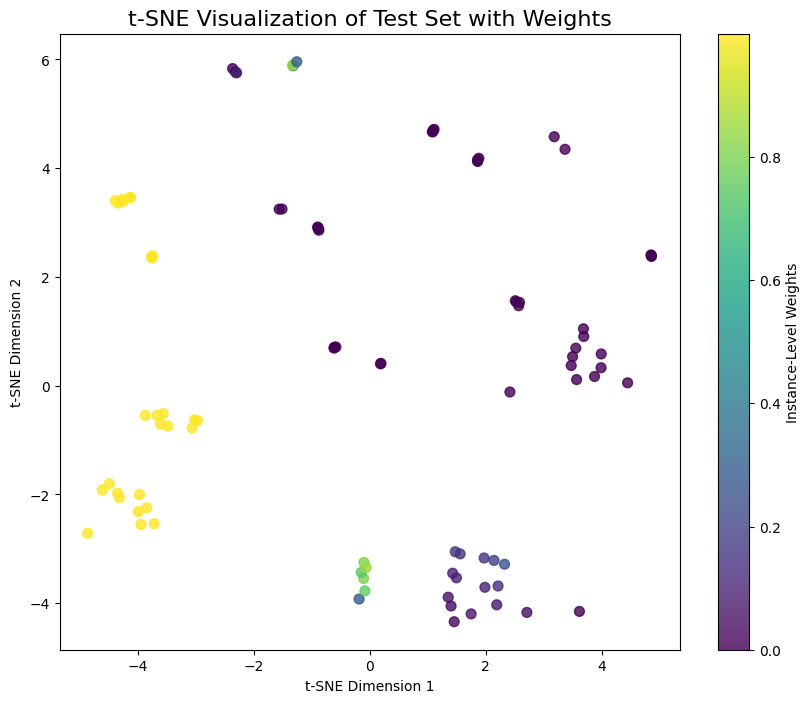}
  \includegraphics[width=0.31\textwidth]{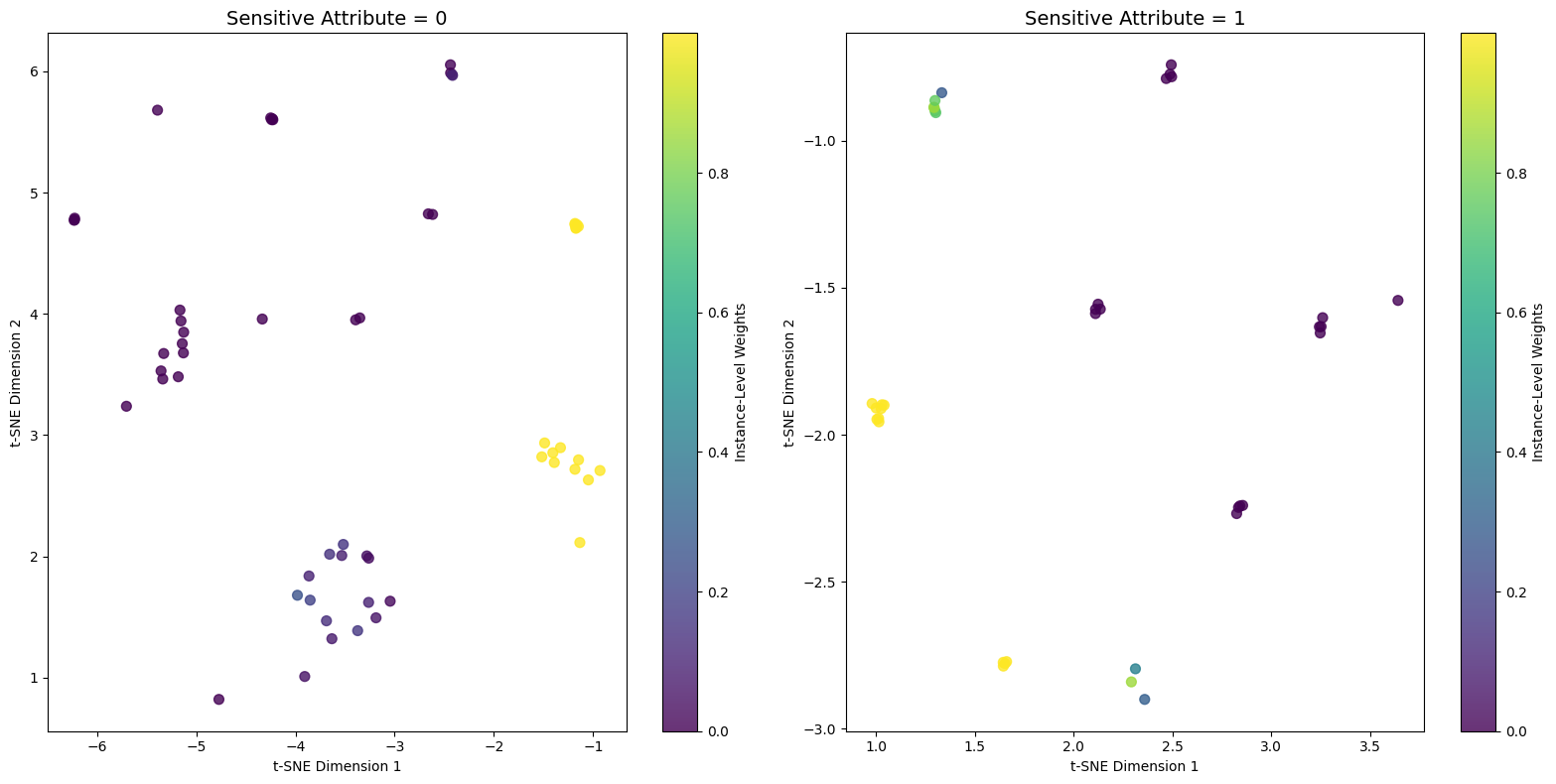}
  \includegraphics[width=0.21\textwidth]{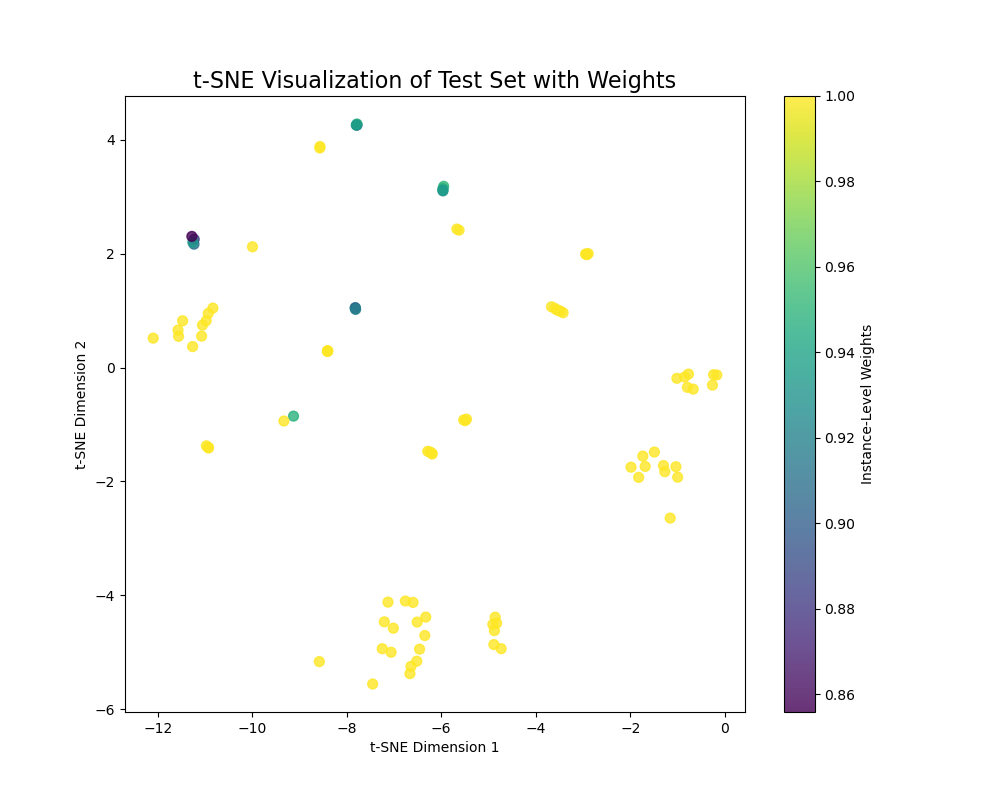}
  \includegraphics[width=0.3\textwidth]{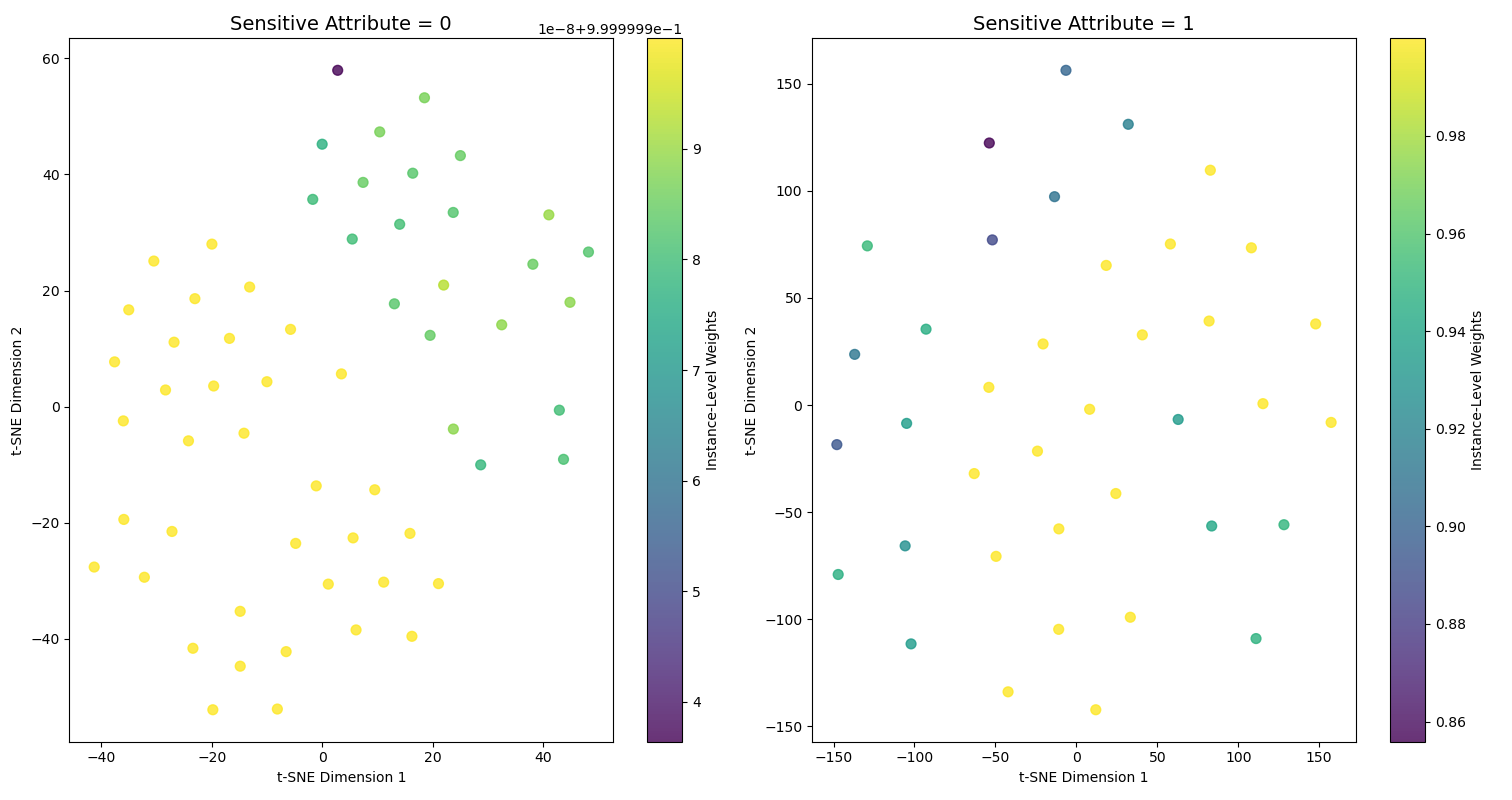}
}
\subfigure[Employee: One/Two Models — Overall, By \texttt{gender}]{
  \includegraphics[width=0.21\textwidth]{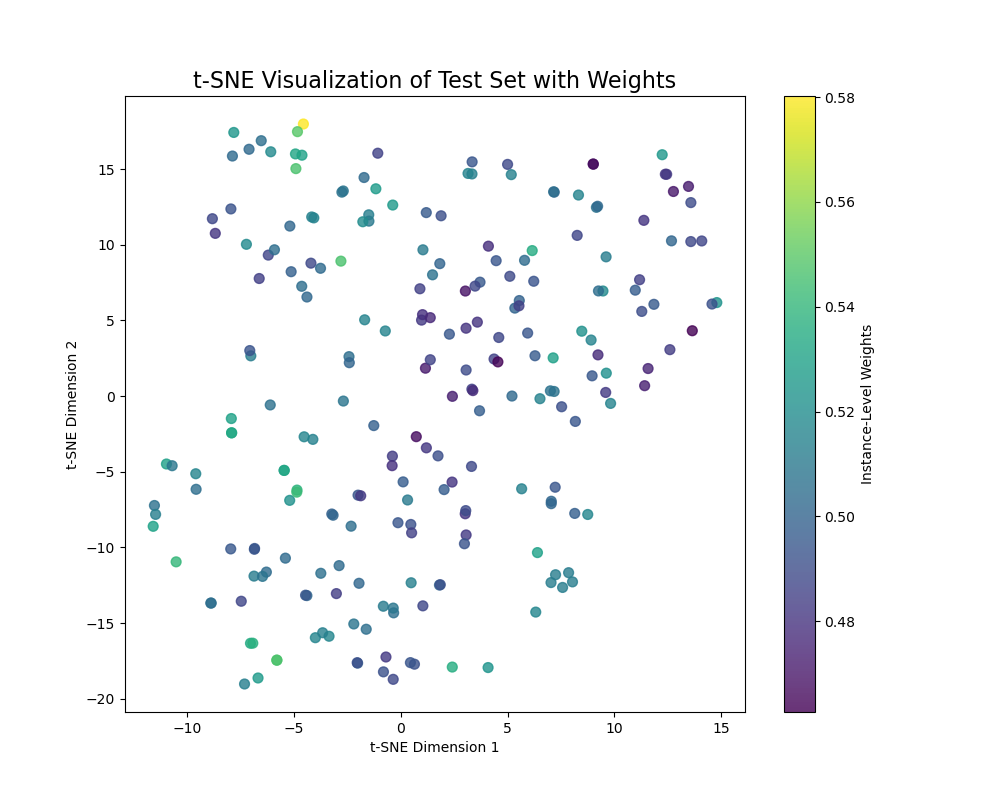}
  \includegraphics[width=0.29\textwidth]{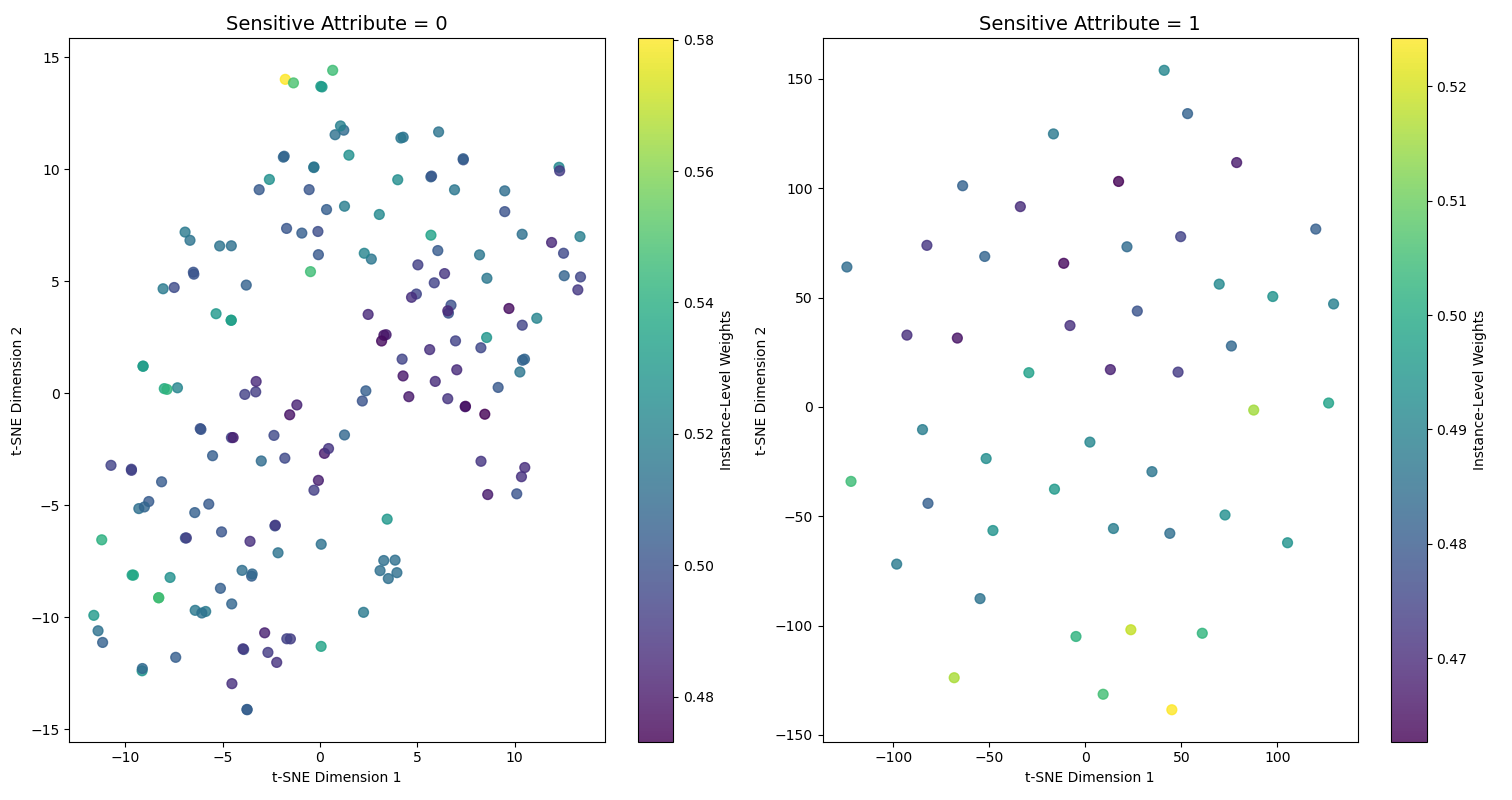}
  \includegraphics[width=0.21\textwidth]{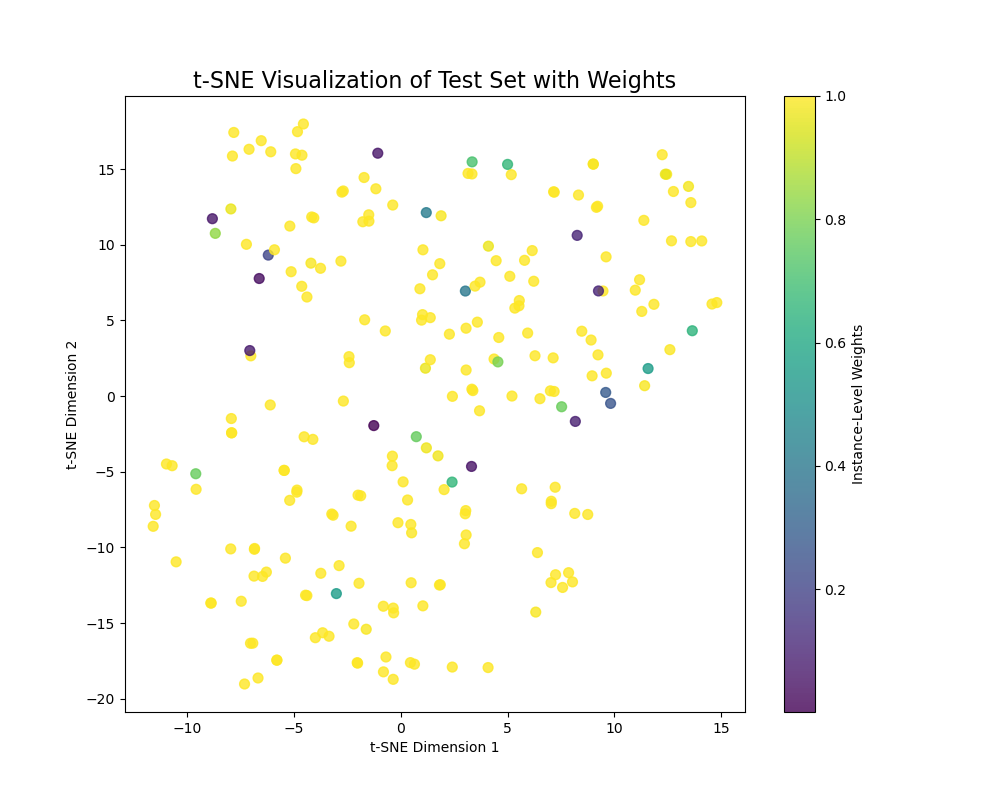}
  \includegraphics[width=0.29\textwidth]{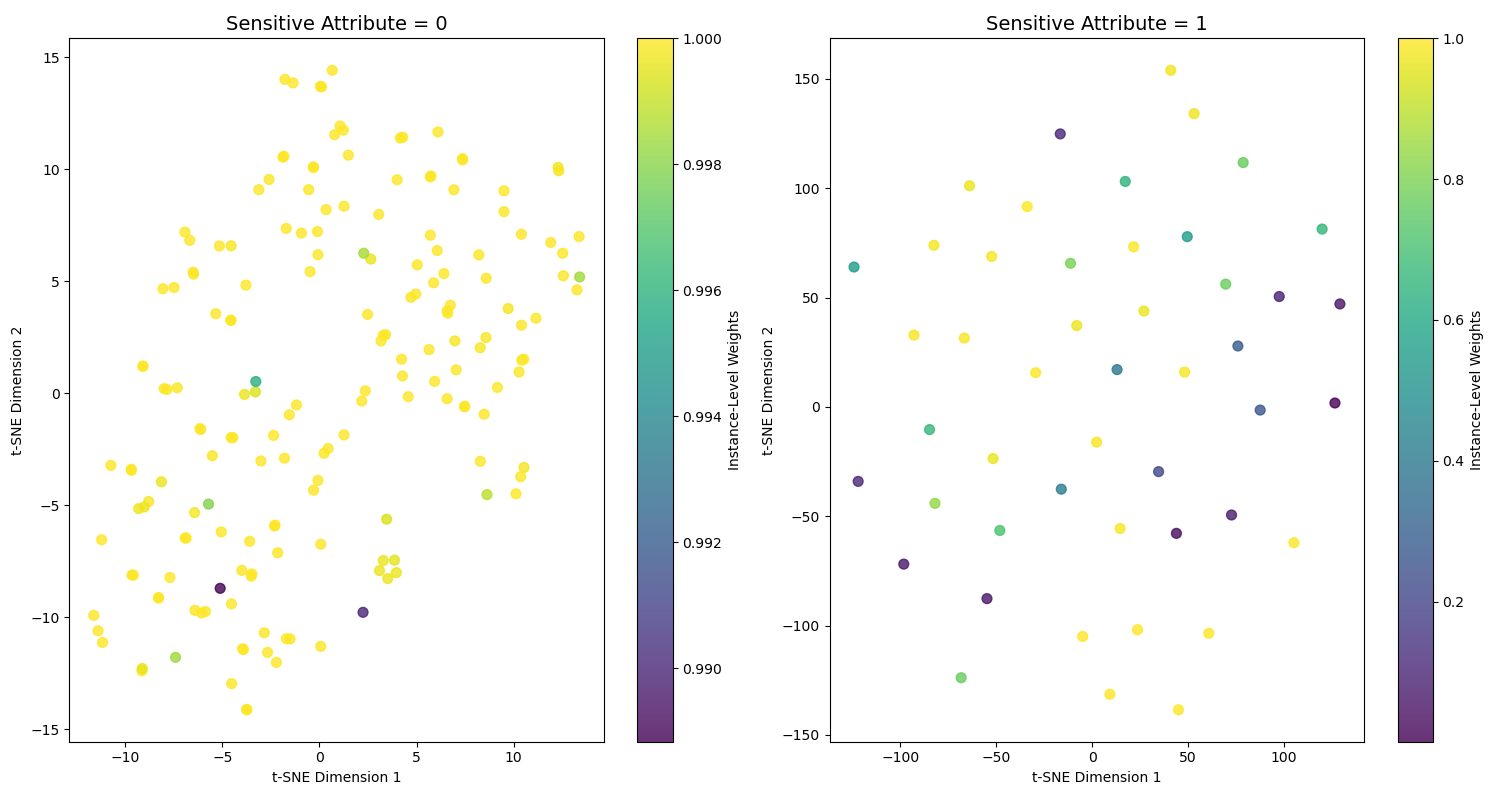}
}
\end{minipage}
}

\caption{\footnotesize
t-SNE visualizations of instance-level weights from MoE and post-processed MoE models across five datasets under single-sensitive (gender) settings. Random Forest and Random Survival Forest are used as base predictors.
}
\label{fig:tsne_others}
\end{figure}

We use t-SNE \cite{vanDerMaaten2008visualizing} to project the instance-level soft assignment weights produced by the \texttt{MoE} gating network into a two-dimensional space. Figure~\ref{fig:tsne_adult_compas} presents these projections for the \textit{Adult} and \textit{COMPAS} datasets under both single-sensitive (i.e., \texttt{gender}) and multi-sensitive (i.e., \texttt{gender + race}) settings, while Figure~\ref{fig:tsne_others} shows the results for the remaining datasets under single-sensitive settings. In each case, we visualize both the overall distribution of instance weights and subgroup-specific patterns. The resulting plots show heterogeneous weight-distribution patterns across observations and, in some cases, visible differences across sensitive groups. These patterns are also present in the multi-sensitive settings and across both the one-pre-trained and two-pre-trained MoE configurations. The visualizations therefore provide a qualitative illustration of variation in the learned instance-specific weights, rather than evidence of a particular group-specific routing mechanism.

\section{Discussion and Limitations}
\label{sec:discussion}

The experimental results show that the proposed framework can produce a range of performance--fairness operating points by varying the regularization parameter \(\lambda\). The relative behaviour of the Mixture and MoE variants depends on the dataset and model configuration. The Mixture formulation offers a parsimonious and interpretable global adjustment, whereas the MoE formulation allows the contribution of the performance-oriented and fairness-oriented components to vary across observations. This additional flexibility can be useful for heterogeneous data, although it does not guarantee a uniformly better trade-off in every setting.

The one- and two-pre-trained formulations address different deployment scenarios. The one-pre-trained setting is applicable when only an existing performance-oriented model is available, while the two-pre-trained setting provides a modular option when a separate fairness-oriented model has already been constructed. In both cases, the performance-oriented base model remains fixed, supporting the use of black-box predictors provided that their prediction outputs are available. Additional experiments with deeper three-layer MLP architectures on the \textit{Adult} and \textit{COMPAS} datasets further indicate that the observed trade-offs are not restricted to the shallow MLP architecture used in the main experiments.

Several limitations should be noted. The empirical evaluation is restricted to binary classification, regression, and survival analysis under the task-specific group-fairness criteria considered in this study, and the observed behaviour may not generalise to other tasks, fairness definitions, datasets, or deployment settings. Moreover, the framework characterises rather than removes the inherent trade-off between predictive performance and group fairness, so the appropriate operating point remains application-dependent.

\section{Conclusion and Future Work}
In this work, we introduced an ensemble-based post-processing framework for fairness-aware prediction. It combines a performance-optimized model, used alone or with a separately pre-trained fairness-oriented model, with a simpler model that adjusts the resulting predictions under an explicit fairness regularization objective. Our approach supports both mixture model and mixture-of-experts weighting strategies, is instantiated for classification, regression, and survival analysis, and does not require access to sensitive attributes during inference. Importantly, the framework operates independently of the internal structure of the base predictor, provided that its predictions are accessible.

Future work could explore extending the framework to incorporate additional ensemble strategies and alternative constructions of the fairness-oriented component, with the aim of further improving flexibility across different datasets and fairness objectives. One promising direction is to expand the mixture-of-experts approach by including multiple auxiliary models with distinct inductive biases or fairness objectives, which may improve flexibility across heterogeneous data distributions. Investigating whether the proposed approach can be extended to settings such as individual fairness is an important direction for future work. Additionally, validating the framework across a broader range of domains, including real-time decision systems and high-stakes areas such as healthcare and finance, would help assess the generalizability and robustness of the proposed approach.

\backmatter

\newpage
\bibliography{sn-bibliography}% common bib file

@inproceedings{tifreafrappe,
  title={FRAPP{\'E}: A Group Fairness Framework for Post-Processing Everything},
  author={Tifrea, Alexandru and Lahoti, Preethi and Packer, Ben and Halpern, Yoni and Beirami, Ahmad and Prost, Flavien},
  year={2024},
  booktitle={Forty-first International Conference on Machine Learning}
}

@inproceedings{zhao2023fairness,
  title={Fairness-aware processing techniques in survival analysis: Promoting equitable predictions},
  author={Zhao, Zhouting and Ng, Tin Lok James},
  booktitle={Joint European Conference on Machine Learning and Knowledge Discovery in Databases},
  pages={460--476},
  year={2023},
  organization={Springer}
}

@article{filippi2023local,
  title={Local Law 144: A Critical Analysis of Regression Metrics},
  author={Filippi, Giulio and Zannone, Sara and Hilliard, Airlie and Koshiyama, Adriano},
  journal={ arXiv:2302.04119},
  year={2023}
}

@incollection{angwin2022machine,
  title={Machine bias},
  author={Angwin, Julia and Larson, Jeff and Mattu, Surya and Kirchner, Lauren},
  booktitle={Ethics of data and analytics},
  pages={254--264},
  year={2022},
  publisher={Auerbach Publications},
address={Florida, USA}
}

@article{breiman2001random,
  title={Random forests},
  author={Breiman, Leo},
  journal={Machine learning},
  volume={45},
  pages={5--32},
  year={2001},
  publisher={Springer}
}

@article{segal2004machine,
  title={Machine learning benchmarks and random forest regression},
  author={Segal, Mark R},
  year={2004}
}

@article{ishwaran2008random,
  title={Random survival forests},
  author={Ishwaran, Hemant and Kogalur, Udaya B and Blackstone, Eugene H and Lauer, Michael S},
  year={2008}
}

@article{hosmer1999regression,
  title={Regression modeling of time to event data},
  author={Hosmer, David W and Lemeshow, Stanley and May, Susanne},
  journal={New York},
  year={1999}
}

@misc{wijaya2020employee,
  author       = {Wijaya, D.},
  title        = {Employee Turnover Dataset [Data set]},
  year         = {2020},
  howpublished = {\url{https://www.kaggle.com/datasets/davinwijaya/employee-turnover}},
  note         = {Accessed: 2025-08-02}
}

@article{wightman1998lsac,
  title={LSAC National Longitudinal Bar Passage Study. LSAC Research Report Series.},
  author={Wightman, Linda F},
  year={1998},
  publisher={ERIC}
}

@inproceedings{zafar2017fairness,
  title={Fairness constraints: Mechanisms for fair classification},
  author={Zafar, Muhammad Bilal and Valera, Isabel and Rogriguez, Manuel Gomez and Gummadi, Krishna P},
  booktitle={Artificial intelligence and statistics},
  pages={962--970},
  year={2017},
  organization={PMLR}
}

@inproceedings{dwork2012fairness,
  title={Fairness through awareness},
  author={Dwork, Cynthia and Hardt, Moritz and Pitassi, Toniann and Reingold, Omer and Zemel, Richard},
  booktitle={Proceedings of the 3rd innovations in theoretical computer science conference},
  pages={214--226},
  year={2012}
}

@article{berk2017convex,
  title={A convex framework for fair regression},
  author={Berk, Richard and Heidari, Hoda and Jabbari, Shahin and Joseph, Matthew and Kearns, Michael and Morgenstern, Jamie and Neel, Seth and Roth, Aaron},
  journal={ arXiv:1706.02409},
  year={2017}
}

@article{kleindessner2020notion,
  title={A notion of individual fairness for clustering},
  author={Kleindessner, Matth{\"a}us and Awasthi, Pranjal and Morgenstern, Jamie},
  journal={ arXiv:2006.04960},
  year={2020}
}

@inproceedings{li2021approximate,
  title={Approximate group fairness for clustering},
  author={Li, Bo and Li, Lijun and Sun, Ankang and Wang, Chenhao and Wang, Yingfan},
  booktitle={International conference on machine learning},
  pages={6381--6391},
  year={2021},
  organization={PMLR}
}

@article{sonabend2022flexible,
  title={Flexible group fairness metrics for survival analysis},
  author={Sonabend, Raphael and Pfisterer, Florian and Mishler, Alan and Schauer, Moritz and Burk, Lukas and Mukherjee, Sumantrak and Vollmer, Sebastian},
  journal={ arXiv:2206.03256},
  year={2022}
}

@inproceedings{keya2021equitable,
  title={Equitable allocation of healthcare resources with fair survival models},
  author={Keya, Kamrun Naher and Islam, Rashidul and Pan, Shimei and Stockwell, Ian and Foulds, James},
  booktitle={Proceedings of the 2021 siam international conference on data mining (sdm)},
  pages={190--198},
  year={2021},
  organization={SIAM}
}

@inproceedings{rahman2022fair,
  title={Fair and interpretable models for survival analysis},
  author={Rahman, Md Mahmudur and Purushotham, Sanjay},
  booktitle={Proceedings of the 28th ACM SIGKDD Conference on Knowledge Discovery and Data Mining},
  pages={1452--1462},
  year={2022}
}

@article{pessach2022review,
  title={A review on fairness in machine learning},
  author={Pessach, Dana and Shmueli, Erez},
  journal={ACM Computing Surveys (CSUR)},
  volume={55},
  number={3},
  pages={1--44},
  year={2022},
  publisher={ACM New York, NY}
}

@article{mehrabi2021survey,
  title={A survey on bias and fairness in machine learning},
  author={Mehrabi, Ninareh and Morstatter, Fred and Saxena, Nripsuta and Lerman, Kristina and Galstyan, Aram},
  journal={ACM computing surveys (CSUR)},
  volume={54},
  number={6},
  pages={1--35},
  year={2021},
  publisher={ACM New York, NY, USA}
}

@article{caton2024fairness,
  title={Fairness in machine learning: A survey},
  author={Caton, Simon and Haas, Christian},
  journal={ACM Computing Surveys},
  volume={56},
  number={7},
  pages={1--38},
  year={2024},
  publisher={ACM New York, NY}
}

@article{kamiran2012data,
  title={Data preprocessing techniques for classification without discrimination},
  author={Kamiran, Faisal and Calders, Toon},
  journal={Knowledge and information systems},
  volume={33},
  number={1},
  pages={1--33},
  year={2012},
  publisher={Springer}
}

@inproceedings{feldman2015certifying,
  title={Certifying and removing disparate impact},
  author={Feldman, Michael and Friedler, Sorelle A and Moeller, John and Scheidegger, Carlos and Venkatasubramanian, Suresh},
  booktitle={proceedings of the 21th ACM SIGKDD international conference on knowledge discovery and data mining},
  pages={259--268},
  year={2015}
}

@inproceedings{zhang2018mitigating,
  title={Mitigating unwanted biases with adversarial learning},
  author={Zhang, Brian Hu and Lemoine, Blake and Mitchell, Margaret},
  booktitle={Proceedings of the 2018 AAAI/ACM Conference on AI, Ethics, and Society},
  pages={335--340},
  year={2018}
}

@article{pleiss2017fairness,
  title={On fairness and calibration},
  author={Pleiss, Geoff and Raghavan, Manish and Wu, Felix and Kleinberg, Jon and Weinberger, Kilian Q},
  journal={Advances in neural information processing systems},
  volume={30},
  year={2017}
}

@article{jacobs1991adaptive,
  title={Adaptive mixtures of local experts},
  author={Jacobs, Robert A and Jordan, Michael I and Nowlan, Steven J and Hinton, Geoffrey E},
  journal={Neural computation},
  volume={3},
  number={1},
  pages={79--87},
  year={1991},
  publisher={MIT Press}
}

@inproceedings{calders2009building,
  title={Building classifiers with independency constraints},
  author={Calders, Toon and Kamiran, Faisal and Pechenizkiy, Mykola},
  booktitle={2009 IEEE international conference on data mining workshops},
  pages={13--18},
  year={2009},
}

@article{zhu1997algorithm,
  title={Algorithm 778: L-BFGS-B: Fortran subroutines for large-scale bound-constrained optimization},
  author={Zhu, Ciyou and Byrd, Richard H and Lu, Peihuang and Nocedal, Jorge},
  journal={ACM Transactions on mathematical software (TOMS)},
  volume={23},
  number={4},
  pages={550--560},
  year={1997},
  publisher={ACM New York, NY, USA}
}

@article{chierichetti2017fair,
  title={Fair clustering through fairlets},
  author={Chierichetti, Flavio and Kumar, Ravi and Lattanzi, Silvio and Vassilvitskii, Sergei},
  journal={Advances in neural information processing systems},
  volume={30},
  year={2017}
}

@article{han2023ffb,
  title={FFB: A Fair Fairness Benchmark for In-Processing Group Fairness Methods},
  author={Han, Xiaotian and Chi, Jianfeng and Chen, Yu and Wang, Qifan and Zhao, Han and Zou, Na and Hu, Xia},
  journal={ arXiv:2306.09468},
  year={2023}
}

@article{petersen2021post,
  title={Post-processing for individual fairness},
  author={Petersen, Felix and Mukherjee, Debarghya and Sun, Yuekai and Yurochkin, Mikhail},
  journal={Advances in Neural Information Processing Systems},
  volume={34},
  pages={25944--25955},
  year={2021}
}

@article{di2024post,
  title={Post-processing fairness with minimal changes},
  author={Di Gennaro, Federico and Laugel, Thibault and Grari, Vincent and Renard, Xavier and Detyniecki, Marcin},
  journal={arXiv preprint arXiv:2408.15096},
  year={2024}
}

@inproceedings{xian2023fair,
  title={Fair and optimal classification via post-processing},
  author={Xian, Ruicheng and Yin, Lang and Zhao, Han},
  booktitle={International Conference on Machine Learning},
  pages={37977--38012},
  year={2023},
}

@inproceedings{jiang2020wasserstein,
  title={Wasserstein fair classification},
  author={Jiang, Ray and Pacchiano, Aldo and Stepleton, Tom and Jiang, Heinrich and Chiappa, Silvia},
  booktitle={Uncertainty in artificial intelligence},
  pages={862--872},
  year={2020},
  organization={PMLR}
}

@article{alghamdi2022beyond,
  title={Beyond adult and compas: Fairness in multi-class prediction},
  author={Alghamdi, Wael and Hsu, Hsiang and Jeong, Haewon and Wang, Hao and Michalak, P Winston and Asoodeh, Shahab and Calmon, Flavio P},
  journal={arXiv:2206.07801},
  year={2022}
}

@inproceedings{kamishima2012fairness,
  title={Fairness-aware classifier with prejudice remover regularizer},
  author={Kamishima, Toshihiro and Akaho, Shotaro and Asoh, Hideki and Sakuma, Jun},
  booktitle={Machine Learning and Knowledge Discovery in Databases: European Conference,  Proceedings, Part II 23},
  pages={35--50},
  year={2012},
  organization={Springer}
}

@inproceedings{agarwal2019fair,
  title={Fair regression: Quantitative definitions and reduction-based algorithms},
  author={Agarwal, Alekh and Dud{\'\i}k, Miroslav and Wu, Zhiwei Steven},
  booktitle={International Conference on Machine Learning},
  pages={120--129},
  year={2019},
  organization={PMLR}
}

@article{vanDerMaaten2008visualizing,
  title={Visualizing data using t-SNE},
  author={van der Maaten, Laurens and Hinton, Geoffrey},
  journal={Journal of Machine Learning Research},
  volume={9},
  number={Nov},
  pages={2579--2605},
  year={2008}
}

@misc{insurance_dataset,
  author       = {Noordeen, M.},
  title        = {Insurance Premium Prediction [Data set]},
  year         = {2020},
  howpublished = {\url{https://www.kaggle.com/datasets/noordeen/insurance-premium-prediction}},
  note         = {Accessed: 2025-08-02. Originally compiled from Prof. Eric Suess's 2017 course materials.}
}

@article{becker1996adult,
  title={Adult},
  author={Becker, Barry and Kohavi, Ronny},
  journal={UCI Machine Learning Repository},
  volume={10},
  pages={C5XW20},
  year={1996}
}

@misc{german_credit_data,
  author       = {Hofmann, Hans},
  title        = {{Statlog (German Credit Data)}},
  year         = {1994},
  howpublished = {UCI Machine Learning Repository},
  note         = {{DOI}: https://doi.org/10.24432/C5NC77}
}

@article{rumelhart1986learning,
  title={Learning representations by back-propagating errors},
  author={Rumelhart, David E and Hinton, Geoffrey E and Williams, Ronald J},
  journal={nature},
  volume={323},
  number={6088},
  pages={533--536},
  year={1986},
  publisher={Nature Publishing Group UK London}
}

@inproceedings{kamiran2012decision,
  title={Decision theory for discrimination-aware classification},
  author={Kamiran, Faisal and Karim, Asim and Zhang, Xiangliang},
  booktitle={2012 IEEE 12th international conference on data mining},
  pages={924--929},
  year={2012},
  organization={IEEE}
}

@article{bellamy2018ai,
  title={AI Fairness 360: An extensible toolkit for detecting, understanding, and mitigating unwanted algorithmic bias. arXivpreprint},
  author={Bellamy, Rachel KE and Dey, Kuntal and Hind, Michael and Hoffman, Samuel C and Houde, Stephanie and Kannan, Kalapriya and Lohia, Pranay and Martino, Jacquelyn and Mehta, Sameep and Mojsilovic, Aleksandra and others},
  journal={arXiv preprint arXiv:1810.01943},
  year={2018}
}

@article{hardt2016equality,
  title={Equality of opportunity in supervised learning},
  author={Hardt, Moritz and Price, Eric and Srebro, Nati},
  journal={Advances in neural information processing systems},
  volume={29},
  year={2016}
}

@article{barocas2020fairness,
  title={Fairness and machine learning},
  author={Barocas, Solon and Hardt, Moritz and Narayanan, Arvind},
  journal={Recommender systems handbook},
  volume={1},
  pages={453--459},
  year={2020},
  publisher={fairmlbook. org}
}

@article{kamishima2024fairmoe,
  title={FairMOE: counterfactually-fair mixture of experts with levels of interpretability},
  author={Kamishima, Toshihiro and Oizumi, Ryohei and Akaho, Shotaro},
  journal={Machine Learning},
  volume={113},
  pages={2535--2564},
  year={2024},
  publisher={Springer}
}

@inproceedings{kearns2018preventing,
  title={Preventing Fairness Gerrymandering: Auditing and Learning for Subgroup Fairness},
  author={Kearns, Michael and Neel, Seth and Roth, Aaron and Wu, Zhiwei Steven},
  booktitle={Proceedings of the 35th International Conference on Machine Learning (ICML)},
  pages={2564--2572},
  year={2018},
  organization={PMLR}
}

@inproceedings{agarwal2018reductions,
  title={A reductions approach to fair classification},
  author={Agarwal, Alekh and Beygelzimer, Alina and Dud{\'\i}k, Miroslav and Langford, John and Wallach, Hanna},
  booktitle={International conference on machine learning},
  pages={60--69},
  year={2018},
  organization={PMLR}
}

@inproceedings{mary2019fairness,
  title={Fairness-aware learning for continuous attributes and treatments},
  author={Mary, J{\'e}r{\'e}mie and Calauzenes, Cl{\'e}ment and El Karoui, Noureddine},
  booktitle={International conference on machine learning},
  pages={4382--4391},
  year={2019},
  organization={PMLR}
}

@article{graf1999assessment,
  title={Assessment and comparison of prognostic classification schemes for survival data},
  author={Graf, Erika and Schmoor, Claudia and Sauerbrei, Willi and Schumacher, Martin},
  journal={Statistics in medicine},
  volume={18},
  number={17-18},
  pages={2529--2545},
  year={1999},
  publisher={Wiley Online Library}
}

@article{uno2007evaluating,
  title={Evaluating prediction rules for t-year survivors with censored regression models},
  author={Uno, Hajime and Cai, Tianxi and Tian, Lu and Wei, Lee-Jen},
  journal={Journal of the American Statistical Association},
  volume={102},
  number={478},
  pages={527--537},
  year={2007},
  publisher={Taylor \& Francis}
}

@article{harrell1982evaluating,
  title={Evaluating the yield of medical tests},
  author={Harrell, Frank E and Califf, Robert M and Pryor, David B and Lee, Kerry L and Rosati, Robert A},
  journal={Jama},
  volume={247},
  number={18},
  pages={2543--2546},
  year={1982},
  publisher={American Medical Association}
}

@article{powell1964efficient,
  title={An efficient method for finding the minimum of a function of several variables without calculating derivatives},
  author={Powell, Michael JD},
  journal={The Computer Journal},
  volume={7},
  number={2},
  pages={155--162},
  year={1964},
  publisher={Oxford University Press}
}

@inproceedings{yang2026bayes,
  title={Bayes-optimal fair classification with multiple sensitive features},
  author={Yang, Yi and Huang, Yinghui and Chang, Xiangyu},
  booktitle={Proceedings of the AAAI Conference on Artificial Intelligence},
  volume={40},
  number={33},
  pages={27693--27701},
  year={2026}
}

@article{yeh2009comparisons,
  title={The comparisons of data mining techniques for the predictive accuracy of probability of default of credit card clients},
  author={Yeh, I-Cheng and Lien, Che-hui},
  journal={Expert Systems with Applications},
  volume={36},
  number={2},
  pages={2473--2480},
  year={2009},
  publisher={Elsevier},
  doi={10.1016/j.eswa.2007.12.020}
}

\end{document}